\documentclass[journal]{IEEEtran}
\usepackage{amsmath,amsfonts}
\usepackage{algorithmic}
\usepackage{algorithm}
\usepackage{array}
\usepackage[caption=false,font=normalsize,labelfont=sf,textfont=sf]{subfig}
\usepackage{textcomp}
\usepackage{amssymb} 
\usepackage{stfloats}
\usepackage{placeins}
\usepackage{url}

\usepackage{verbatim}
\usepackage{graphicx}
\usepackage{cite}
\usepackage{booktabs} 
\usepackage{multirow} 
\usepackage{float} 
\usepackage{hyperref}
\usepackage{makecell}

\begin{document}

\title{MISApp: Multi-Hop Intent-Aware Session Graph Learning for Next App Prediction}
    
    

    

\author{
    Yunchi~Yang$^{\dagger}$, 
    Longlong~Li$^{\dagger}$, 
    Jianliang~Wu, 
    and~Cunquan~Qu$^{*}$
    
    \thanks{$^{\dagger}$Yunchi Yang and Longlong Li contributed equally to this work.}
    
    \thanks{Yunchi Yang is with the Data Science Institute, Shandong University, Jinan 250100, China (e-mail: ycyang@mail.sdu.edu.cn).}
    
    \thanks{Longlong Li is with the School of Physical and Mathematical Sciences, Nanyang Technological University, Singapore 637371, Singapore (e-mail: longlong.li@ntu.edu.sg).}

    \thanks{Jianliang Wu is with the School of Mathematics, Shandong University, Jinan 250100, China (e-mail: jlwu@sdu.edu.cn).}
    
    \thanks{$^{*}$Cunquan Qu is the corresponding author and is with the Data Science Institute, Shandong University, Jinan 250100, China (e-mail: cqqu@sdu.edu.cn).}
}

\maketitle

\begin{abstract}
Predicting the next mobile app a user will launch is essential for proactive mobile services. Yet accurate prediction remains challenging in real-world settings, where user intent can shift rapidly within short sessions and user-specific historical profiles are often sparse or unavailable, especially under cold-start conditions. Existing approaches mainly model app usage as sequential behavior or local session transitions, limiting their ability to capture higher-order structural dependencies and evolving session intent. To address this issue, we propose MISApp, a profile-free framework for next app prediction based on multi-hop session graph learning. MISApp constructs multi-hop session graphs to capture transition dependencies at different structural ranges, learns session representations through lightweight graph propagation, incorporates temporal context and similarity-based spatial categorization to characterize session conditions, and captures intent evolution from recent interactions. Experiments on two real-world app usage datasets show that MISApp consistently outperforms competitive baselines under both standard and cold-start settings, while maintaining a favorable balance between predictive accuracy and practical efficiency. Further analyses show that multi-hop relations capture higher-hop-specific predictive signals beyond direct 1-Hop adjacency, and that the learned hop-level attention weights align well with structural relevance, providing both empirical and interpretable evidence for the effectiveness of the proposed multi-hop modeling strategy.
\end{abstract}

\section{Introduction}
With the widespread adoption of smartphones, mobile applications have become an integral part of daily digital life~\cite{c2}. Modern app ecosystems span diverse categories~\cite{ouyang2018modeling, chu2021air, gruning2023directing, zhu2013mobile}, including social media, online shopping, navigation, short-video platforms, and productivity tools, enabling users to conveniently access digital services in a wide range of scenarios~\cite{zhang2020app, zhu2013mobile}. In this context, accurately predicting the next app a user is likely to launch is important for proactive mobile services, as it can improve app recommendation quality, optimize device-level resource allocation~\cite{chen2017powerful, oliner2013carat}, enhance system responsiveness~\cite{xu2016understanding}, and facilitate network-level scheduling. Consequently, next app prediction has become an important problem in mobile user behavior modeling with both practical and research significance~\cite{liu2017effective, li2022smartphone}.

Despite its importance, accurate next app prediction remains challenging in real-world settings~\cite{9831540}. First, user intent may change rapidly within short sessions, and such changes are often context-dependent rather than strictly sequential. As a result, models that mainly rely on fixed-order dependencies, such as RNN-based approaches~\cite{xu2019recurrent} and Transformer-based sequence models~\cite{vaswani2017attention,li2025transformer}, may not fully capture fine-grained intent evolution from limited interactions. This challenge becomes more pronounced in cold-start scenarios, where user-specific histories are sparse or unavailable and intent must be inferred directly from current session behaviors.

Second, session-level app usage contains not only immediate transitions between consecutive apps but also higher-order structural dependencies that reflect latent behavioral routines and semantically related app usages~\cite{zhao2019appusage2vec, chen2019cap}. Existing methods that  mainly focus on raw sequences or 1-Hop session graphs provide only a limited view of session structure, which may overlook long-range dependencies and cross-app latent relations.

Third, session-level app usage is also shaped by contextual conditions such as time and location~\cite{shin2012understanding, baeza2015predicting}. These cues can modulate the meaning of transition patterns and provide useful evidence for intent inference~\cite{aliannejadi2021context, sun2025appformer}. However, many existing methods incorporate contextual information only weakly or independently from session structure, limiting their ability to jointly reason over behavioral dependencies and session conditions.

To address these challenges, we propose MISApp, a profile-free framework that infers next-app intent directly from session-level behaviors without assuming the availability of static user profiles. MISApp captures multi-range transition dependencies through multi-hop session graphs, models intent evolution from recent interactions, and incorporates temporal and spatial context to characterize session conditions. By combining structural dependency modeling with session-level intent inference, MISApp supports robust and accurate next app prediction, particularly in cold-start scenarios where personalized historical information is limited.

The contributions of our work are summarized as follows:
\begin{itemize}
\item 
We propose a multi-hop session graph framework that explicitly decomposes app transitions into multiple structural ranges. By modeling both direct and higher-order dependencies within sessions, MISApp captures richer behavioral structure than conventional sequential or 1-Hop graph representations.

\item 
We develop a profile-free intent modeling framework that infers evolving user intent directly from recent session behaviors, without relying on static user profiles or long-term historical attributes. This design improves robustness in cold-start scenarios and better reflects short-term intent dynamics.

\item
Extensive experiments on two real-world datasets show that MISApp outperforms competitive baselines under both standard and cold-start settings, while maintaining favorable practical efficiency. Additional analyses indicate that the learned hop-level attention aligns well with structural relevance, supporting the structural interpretability of the proposed multi-hop modeling mechanism.
\end{itemize}

\section{RELATED WORK}
Modeling mobile user behavior for the next app prediction fundamentally requires capturing structural dependency patterns within sessions, where different transition ranges may encode distinct semantic roles. This section reviews three research domains most relevant to our work: mobile app usage prediction, graph neural networks, and Transformer-based time-series predictors.

Mobile app usage prediction aims to infer the next app a user is most likely to launch based on their historical behaviors and contextual cues. Prior studies typically cast this task as a time-series prediction problem, emphasizing dependencies across consecutive app launches. Classical  models such as MRU and MFU~\cite{shin2012understanding} offer simple baselines but overlook temporal and contextual dependencies. To better capture sequential dynamics, deep learning–based models such as DeepApp~\cite{xia2020deepapp}, DeepPattern~\cite{suleiman2021DeepPatterns}, NeuSA~\cite{aliannejadi2021context}, MAPLE~\cite{khaokaew2024maple}, and AppUsage2Vec~\cite{zhao2019appusage2vec} aim to encode app usage patterns together with contextual features. However, they struggle to distinguish higher-order behavioral routines, limiting their ability to adapt to rapid intent shifts.

Graph Neural Networks (GNNs) offer a powerful framework for modeling relational structures by propagating information across neighboring nodes~\cite{pareja2020evolvegcn, sankar2020dysat, xu2020inductive}, and thus have been widely adopted in sequential prediction and app recommendation. Graph-based models—such as SA-GCN~\cite{yu2020semantic}, DUGN~\cite{ouyang2022learning}, and various Hypergraph-based approaches~\cite{huang2025predicting, wang2020next}—capture structural dependencies within sessions and effectively model local interaction patterns among apps. However, such designs implicitly entangle structural ranges, as higher-order signals are mixed through repeated propagation rather than explicitly separated. Consequently,  their interpretability remains somewhat limited due to the opaque mixing of structural ranges.

The Transformer~\cite{vaswani2017attention} architecture, built upon self-attention, has demonstrated remarkable effectiveness in capturing long-range dependencies in sequential data. Although recent time-series forecasting architectures such as   Reformer~\cite{kitaev2020reformer}, FEDformer~\cite{zhou2022fedformer}, and TimesNet~\cite{wu2022timesnet} achieve high performance in general time-series forecasting tasks, they are not tailored for user behavior modeling and overlook essential contextual signals such as temporal and spatial factors, which are critical for personalized app prediction. Appformer~\cite{sun2025appformer} integrates multi-modal progressive fusion and sophisticated feature extractors. However, its focus on long-range sequential dependencies limits its ability to capture rapid intention shifts within user-app interactions.

\section{METHODOLOGY}
Let $\mathcal{A}=\{a_1,a_2,\dots,a_{|\mathcal{A}|}\}$ denote the set of all mobile applications, and let $\mathcal{R}=\{r_1,r_2,\dots,r_{|\mathcal{R}|}\}$ denote the set of all base station IDs.
For each user, the raw behavior log is a time-ordered sequence of app usage events
\begin{equation}
\mathcal{X}=\big((a_1,t_1,r_1),(a_2,t_2,r_2),\dots,(a_N,t_N,r_N)\big),
\end{equation}
where $a_i\in\mathcal{A}$ is the app launched at the $i$-th event, $t_i$ is its timestamp, and $r_i\in\mathcal{R}$ is the corresponding base station ID.

We segment the event sequence into sessions according to an inactivity threshold $\delta_t$: two consecutive events are assigned to the same session if and only if
\begin{equation}
t_{i+1}-t_i \le \delta_t.
\end{equation}
In our experiments, $\delta_t=300$ seconds.

After segmentation, each session $S$ is represented as an ordered sequence of app usage events:
\begin{equation}
S=(s_1,s_2,\dots,s_T), \qquad s_i = (a_i,t_i,r_i),
\end{equation}
where $T$ denotes the session window size. To characterize the current context of the session, we use the temporal and spatial information associated with the last app interaction~\cite{li2023you,li2021finding}. Specifically, let $\tau\in\{0,1,\dots,23\}$ denote the hour-of-day of the last app, which represents the temporal context of session $S$, and let $\rho\in\mathcal{R}$ denote the corresponding base station ID, which represents the spatial context of session $S$.

Given the current session $S$ together with its temporal context $\tau$
and spatial context $\rho$, the task is to predict the next app
$a_{T+1}\in\mathcal{A}$. Formally, the model estimates
\begin{equation}
P(a_{T+1}\mid S,\tau,\rho),
\end{equation}
and outputs the most likely app as
\begin{equation}
\hat{a}_{T+1}
=
\arg\max_{a\in\mathcal{A}}
P(a\mid S,\tau,\rho).
\end{equation}
\subsection{Framework Overview}

\begin{figure*}
    \centering
\includegraphics[width=1.0\linewidth]{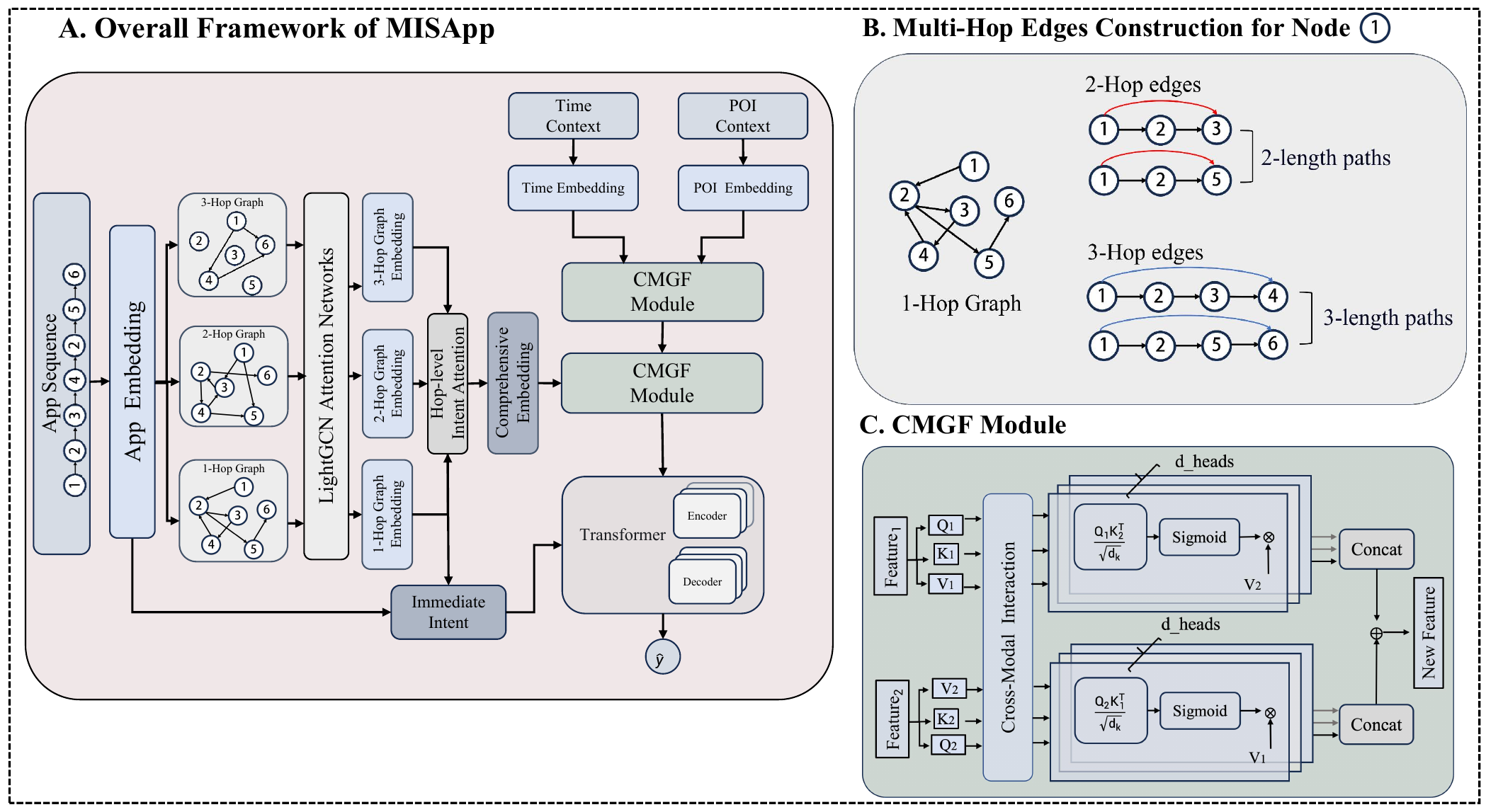}
    \caption{
    Overall architecture of MISApp. The model constructs 1-Hop, 2-Hop, and 3-Hop session graphs and encodes them separately through a LightGCN attention network, followed by hop-level intent attention to obtain a multi-hop structural representation. Temporal and spatial context are incorporated through the Cross-Modal Gated Fusion (CMGF) module. The fused session representation is then combined with the immediate intent derived from recent app interactions and processed by a Transformer encoder–decoder to model intent transitions for the next app prediction.
    }
    \label{fig: model}
\end{figure*}

We propose MISApp, an intent-aware app prediction framework built on multi-hop session graphs. As illustrated in Fig.~\ref{fig: model}A, the architecture comprises four key components:
\begin{itemize}
\item \textbf{Multi-Hop Relation Graph Module}: 
MISApp constructs 1-Hop, 2-Hop, and 3-Hop session graphs based on temporal relations between apps. LightGCN is applied to each graph to learn structural representations.

\item \textbf{Contextual Information Embedding}: 
Session temporal context and spatial context are embedded into a unified vector space, providing complementary signals for modeling user behavior.
\item \textbf{Cross-Modal Gated Fusion}: 
MISApp derives gating weights from cross-modal query–key interactions to regulate the information exchanged between heterogeneous representations.
\item \textbf{Inter-Session Intent Transition Module}: 
To capture evolving user interests, MISApp extracts the most recent apps as an immediate intent signal and leverages a Transformer to model dynamic intent by fusing this signal with session-level structural embeddings.
\end{itemize}



\subsection{Multi-Hop Relation Graph Module}
Each app $a \in \mathcal{A}$ is associated with a learnable embedding vector $\mathbf{e}_A(a) \in \mathbb{R}^d$, where $d$ is the embedding dimension.

The sequential usage behavior of user applications can be effectively modeled as a graph structure to capture the underlying transition patterns within a session. Given a user's historical app interaction sequence in session $S$, we construct a directed session graph:
\begin{equation}
G^S_{1} = (V^S, E^S_1),
\end{equation}
where
\begin{equation}
\begin{split}
V^S &= \{a_t \mid 1 \le t \le T\} \subseteq \mathcal{A}, \\
E^S_1 &= \{(a_t, a_{t+1}) \mid 1 \le t < T\}.
\end{split}
\end{equation}
Each directed edge in $E^S_1$ represents an immediate transition between two consecutive app events.

By representing the session as a directed graph, the model can go beyond raw linear sequences to reflect the intrinsic transition structure of user actions. This 1-Hop session graph $G^S_1$ serves as the fundamental structural representation, providing a basis for capturing more complex, higher-order dependencies in subsequent modules.

Although the 1-Hop session graph structure can effectively capture direct adjacency relations within each session, it cannot model longer-range behavioral dependencies. In session-based app usage, different hop distances encode distinct semantic roles: immediate transitions reflect short-term intent continuity, while higher-order connections capture latent procedural routines. To capture such latent structures, we extend the graph representation to include multi-hop connectivity, as illustrated in Fig.~\ref{fig: model}B.

$E^S_1$ denotes the set of directed edges in the original 1-Hop session graph. 
We construct higher-hop connections by composing edges along reachable paths while excluding self-loops. 
Specifically, the 2-Hop edge set is defined as
\begin{equation}
\begin{split}
E^S_2 = \{ (a_u,a_w) \mid {} & a_u \neq a_w,\ \exists\, a_v \;\text{s.t.} \\
& (a_u,a_v)\in E^S_1,\,(a_v,a_w)\in E^S_1 \}.
\end{split}
\end{equation}
Similarly, the 3-Hop edge set is obtained by extending the composition to three consecutive edges:

\begin{equation}
\begin{split}
E^S_3 = \{ (a_u,a_x) \mid {} & a_u \neq a_x,\ \exists\, a_w \;\text{s.t.} \\
& (a_u,a_w)\in E^S_2,\,(a_w,a_x)\in E^S_1 \}.
\end{split}
\end{equation}
Accordingly, the $\gamma$-Hop session graph is defined as
\begin{equation}
G^S_{\gamma}=\big(V^S,E^S_{\gamma}\big), \qquad \gamma \in\{1,2,3\}.
\end{equation}
The 1-Hop session graph captures immediate transition continuity,
whereas the 2-Hop and 3-Hop session graphs encode longer-range
behavioral dependencies.

The maximum hop number is set to three based on both network-structural intuition and empirical observations. The small-world property of complex networks, commonly illustrated by the ‘six degrees of separation’ phenomenon, is characterized by short average path lengths and high clustering coefficients~\cite{little_world}, indicating that the average shortest-path distance between distinct
nodes is typically small.
For session-based app usage graphs, overly long paths may introduce noisy or weakly related transitions because user sessions are short and intent changes rapidly. Therefore, we focus on 1-Hop, 2-Hop, and 3-Hop relations to capture direct transitions and moderate-range behavioral dependencies.

\begin{figure}[!tbp]
    \centering
    \includegraphics[width=0.75\columnwidth]{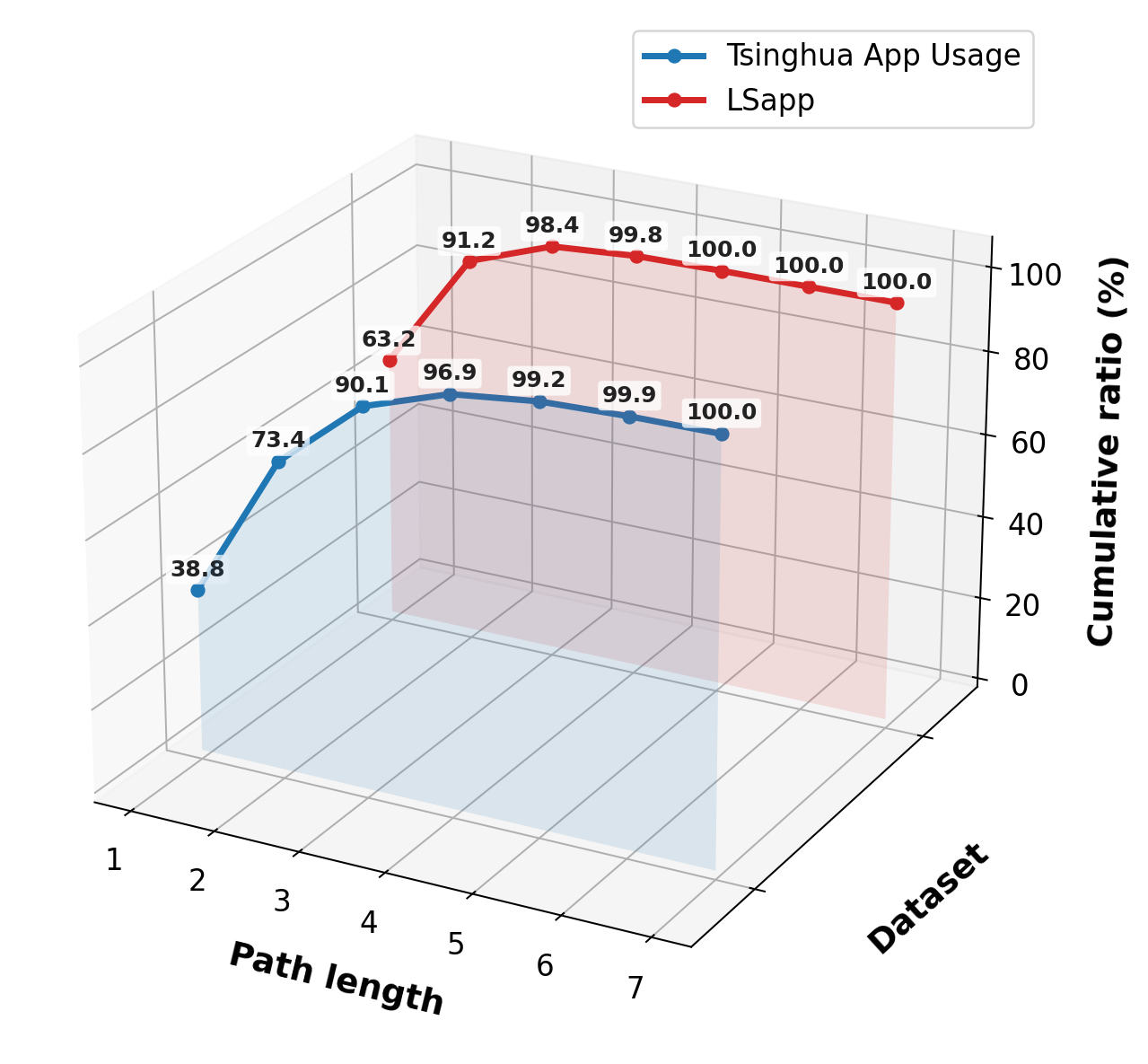}
    \caption{Cumulative distribution of path lengths on the Tsinghua App Usage and LSapp datasets.}
\label{fig:path_distribution}
\end{figure}

To further support this choice, we analyze the normalized path-length distribution over all session samples in the two datasets. As shown in Fig.~\ref{fig:path_distribution}, paths with lengths from 1 to 3 account for 90.1\% of all paths on Tsinghua App Usage and 98.4\% on LSapp, while longer paths occupy only a marginal proportion. This indicates that 3-Hop graphs can cover the dominant structural dependencies while avoiding additional noise and computational overhead from longer-hop connections. A complementary hop-combination experiment in Section~S1 of the supplementary material further shows that the full 1-Hop + 2-Hop + 3-Hop configuration achieves the best overall performance, confirming that these structural ranges provide complementary predictive information.

%



To encode the multi-hop session graphs efficiently, we adopt a LightGCN-based propagation mechanism~\cite{he2020lightgcn}. Compared with conventional graph convolution schemes that combine neighborhood aggregation with feature transformation and nonlinear activation, LightGCN retains a lightweight propagation rule centered on neighbor aggregation. This makes it well suited to our setting, where the goal is to capture structural dependencies over session graphs while keeping the encoder simple and computationally efficient.

To capture structural dependencies at different ranges, we construct three session graphs---1-Hop, 2-Hop, and 3-Hop---and perform LightGCN-based propagation on each graph separately. For the $\gamma$-th graph $G^S_{\gamma}=(V^S, E^S_{\gamma})$, the adjacency relations encode direct transitions, two-step dependencies, and longer-range three-step dependencies, respectively. By applying propagation independently on each graph, MISApp preserves structural information at different hop ranges rather than mixing them within a single propagation process. For the $\gamma$-th graph $G^S_{\gamma}$, the layer-wise propagation is defined as
\begin{equation} 
n_{v}^{(l+1, \gamma)}=
\sum_{u\in \mathcal{N}^{(\gamma)}(v)}
\frac{1}{\sqrt{|\mathcal{N}^{(\gamma)}(v)|}\sqrt{|\mathcal{N}^{(\gamma)}(u)|}}
n_{u}^{(l, \gamma)}, 
\end{equation}
where $n_v^{(0,\gamma)} = \mathbf{e}_{A}(v)$ denotes the initial app embedding in graph $G^S_{\gamma}$, obtained from the App Embedding Layer. $l$ indexes the
propagation layer and $\mathcal{N}^{(\gamma)}(v)$ denotes the set of neighbors of node $v$ in $G^S_{\gamma}$. After $L_g$ layers, node representations are obtained by
layer-wise averaging: 
\begin{equation}
n_{v}^{(\gamma)} = \frac{1}{L_g+1}\sum_{l=0}^{L_g} n_{v}^{(l, \gamma)}.
\end{equation}

To obtain a graph-level embedding for $G^S_{\gamma}$, we apply an attention
mechanism that highlights the apps most relevant to the user's immediate intent.
Specifically, the representation of the last app $a_T$ is used as the query,
while the representations of all nodes in $V^S$ are used as keys. For each
$v\in V^S$, the scaled dot-product score is computed as
\begin{equation}
e_v^{(\gamma)}
=
\frac{
(\mathbf{W}^{Q}\mathbf{n}_{a_T}^{(\gamma)})^\top
(\mathbf{W}^{K}\mathbf{n}_{v}^{(\gamma)})
}{
\sqrt{d}
},
\end{equation}
where $\mathbf{W}^{Q}$ and $\mathbf{W}^{K}$ are learnable projection matrices. The attention
coefficient is then obtained by normalizing the scores over all nodes:
\begin{equation}
\alpha_v^{(\gamma)}
=
\frac{
\exp(e_v^{(\gamma)})
}{
\sum_{u\in V^S}\exp(e_u^{(\gamma)})
}.
\end{equation}
The graph-level embedding is computed as
\begin{equation}
\mathbf{g}^{(\gamma)}
=
\sum_{v\in V^S}
\alpha_v^{(\gamma)}
\mathbf{W}^{V}\mathbf{n}_{v}^{(\gamma)},
\end{equation}
where $\mathbf{W}^{V}$ is the learnable value projection matrix.

Let the immediate intent window size be $K$. The last $K$ applications in the current session are denoted as $\{a_{T-K+1}, \dots, a_T\}$, and their corresponding 1-Hop structural representations are $\{n^{(1)}_{a_{T-K+1}}, \dots, n^{1}_{a_T}\}$. The short-term immediate intent of the current session is represented as:
\begin{equation}
\mathbf{s} = \mathbf{W}[n^{(1)}_{a_{T-K+1}}, ...||..., n^{(1)}_{a_{T}}], 
\end{equation}
where  $\mathbf{W} \in \mathbb{R} ^{d \times Kd}$ are learnable parameters.

To adaptively integrate the information from different graph scales, we propose the Hop-level Intent Attention mechanism. 
We treat the hop-specific representations as structural memory candidates  and utilize the short-term immediate intent $\mathbf{s}$ as a query to dynamically determine the importance of each hop, ensuring that the final session representation is precisely aligned with the user's current interaction context.
The attention weight $W_{hop_{\gamma}}$ for the embedding of the $\gamma$-Hop session graph $\mathbf{g}^{(\gamma)}$ with respect to the intent vector $\mathbf{s}$ is defined as:
\begin{equation}
W_{hop_{\gamma}} = \frac{\exp(\mathbf{s}^{\top} \cdot \mathbf{g}^{(\gamma)})}{\sum_{\gamma=1}^{3} \exp(\mathbf{s}^{\top} \cdot \mathbf{g}^{(\gamma)})},
\end{equation}
then the comprehensive embedding $\mathbf{g}$ is constructed as a weighted sum of the multi-hop session graph embeddings:
\begin{equation}\mathbf{g} = \sum_{\gamma=1}^{3} W_{hop_{\gamma}}\mathbf{g}^{(\gamma)} \in \mathbb{R}^d.
\end{equation}

\subsection{Contextual Information Embedding}
To enrich the session representation with contextual information, MISApp
incorporates two complementary signals: temporal context and spatial context.
The temporal context characterizes when the current session occurs, while the
spatial context captures the functional characteristics of the user's current
location.

\subsubsection{Temporal Context Embedding}

As defined above, we use the hour-of-day $\tau\in\{0,1,\dots,23\}$ associated
with the last app interaction in session $S$ to represent its temporal context.
The corresponding temporal context embedding is defined as
\begin{equation}
\mathbf e^{(\tau)}_{T}
=
\mathbf E_{T}[\tau,:],
\end{equation}
where $\mathbf E_{T}\in\mathbb{R}^{24\times d}$ is a trainable temporal
embedding matrix and $\mathbf e^{(\tau)}_{T}\in\mathbb{R}^{d}$ denotes the
embedding of hour $\tau$.

\subsubsection{Spatial Context Embedding}

Each base station ID in Tsinghua App Usage dataset is associated with a set of POI (Point-of-Interest) environmental attributes, such as the number of nearby pharmacies, hotels, stadiums, and other facilities. Each base station $r\in\mathcal{R}$ is associated with a POI feature vector
\begin{equation}
\mathbf{x}_{r}=(x_{r,1},x_{r,2},\dots,x_{r,M}) \in\mathbb{R}^{M},
\end{equation}
where $M$ denotes the number of POI feature dimensions.

We first perform feature-wise min--max normalization:
\begin{equation}
\tilde{x}_{r,m}
=
\frac{
x_{r,m}-\min_{q\in\mathcal{R}}x_{q,m}
}{
\max_{q\in\mathcal{R}}x_{q,m}-\min_{q\in\mathcal{R}}x_{q,m}
},
\qquad m=1,2,\dots,M.
\end{equation}
Let $\tilde{\mathbf{x}}_{r}\in\mathbb{R}^{M}$ denote the normalized POI vector of base station $r$.
We then define the contextual similarity between two base station IDs $r,q\in\mathcal{R}$ by cosine similarity:
\begin{equation}
\mathrm{sim}_{\mathrm{loc}}(r,q)
=
\frac{
\tilde{\mathbf{x}}_{r}^{\top}\tilde{\mathbf{x}}_{q}
}{
\|\tilde{\mathbf{x}}_{r}\|_{2}\,\|\tilde{\mathbf{x}}_{q}\|_{2}
}.
\end{equation}
Based on these similarities, we construct a sparse base station graph 
$G_{\text{loc}} = (\mathcal{R}, E_{\mathrm{loc}})$, where each base station 
$r \in \mathcal{R}$ is connected to its top-$k_{\mathrm{loc}}$ most similar 
neighbors under $\mathrm{sim}_{\mathrm{loc}}(\cdot,\cdot)$. Formally, the edge 
set is defined as
\begin{equation}
E_{\mathrm{loc}} = \left\{ (r, q) \,\middle|\, 
q \in \mathcal{N}_{k_{\mathrm{loc}}}(r) \right\},
\end{equation}
where $\mathcal{N}_{k_{\mathrm{loc}}}(r)$ denotes the set of top-$k_{\mathrm{loc}}$ 
neighbors of base station $r$.

Let
$
\mathcal{C}=\{C_1,C_2,\dots,C_F\}
$
be the set of connected components of $G_{\mathrm{loc}}$.
Then $\mathcal{C}$ forms a partition of $\mathcal{R}$:
\begin{equation}
C_i\cap C_j=\varnothing \quad (i\neq j),
\qquad
\bigcup_{f=1}^{F} C_f=\mathcal{R}.
\end{equation}
We regard each connected component as a base station category. Accordingly, 
we define the base station category mapping
\begin{equation}
c:\mathcal{R}\to\{1,2,\dots,F\},
\end{equation}
where $c(r)=f$ if and only if $r\in C_f$.

For the current session $S$, We first map $\rho$ to its corresponding category via 
$c(\rho)$. The spatial context embedding 
$\mathbf e^{(\rho)}_{R}\in\mathbb{R}^{d}$ is then defined as
\begin{equation}
\mathbf e^{(\rho)}_{R} = \mathbf E_{R}[c(\rho),:],
\end{equation}
where $\mathbf{E}_{R}\in\mathbb{R}^{F\times d}$ is the learnable embedding 
matrix.

Further analyses of the proposed spatial context construction are provided 
in Section~S3 of the supplementary material, including the sensitivity 
analysis of $k_{\mathrm{loc}}$ and controlled comparisons with alternative 
spatial construction strategies. These results further validate the 
predictive effectiveness and category-separation capability of the proposed 
POI-similarity-based construction.

\subsection{Cross-Modal Gated Fusion}
To effectively integrate heterogeneous yet complementary signals from two distinct modalities, we propose a  Cross-Modal Gated Fusion (CMGF) module, as illustrated in Fig.~\ref{fig: model}C. Each modality is represented by a feature vector $\mathbf{x}_1, \mathbf{x}_2 \in \mathbb{R}^{d}$. Unlike typical static fusion schemes, CMGF enables dynamic, content-aware information exchange through a multi-head bilinear interaction mechanism.

\paragraph{Cross-modal interaction}
For each head $b \in \{1, \dots, B\}$, we first project the input features into query, key, and value subspaces:
\begin{equation}
\begin{aligned}
\mathbf{q}_1^{(b)} &= \mathbf{W}^{Q(b)}_1 \mathbf{x}_1, &
\mathbf{k}_2^{(b)} &= \mathbf{W}^{K(b)}_2 \mathbf{x}_2, &
\mathbf{v}_2^{(b)} &= \mathbf{W}^{V(b)}_2 \mathbf{x}_2, \\
\mathbf{q}_2^{(b)} &= \mathbf{W}^{Q(b)}_2 \mathbf{x}_2, &
\mathbf{k}_1^{(b)} &= \mathbf{W}^{K(b)}_1 \mathbf{x}_1, &
\mathbf{v}_1^{(b)} &= \mathbf{W}^{V(b)}_1 \mathbf{x}_1,
\end{aligned}
\end{equation}
where $\mathbf{W}^{Q(b)}_j, \mathbf{W}^{K(b)}_j, \mathbf{W}^{V(b)}_j \in \mathbb{R}^{d_b \times d}$ are learnable projection matrices. The compatibility between modalities is then measured via scaled dot products between cross-modal query-key pairs:
\begin{equation}
\begin{aligned}
g_{1 \leftarrow 2}^{(b)} &= \sigma\!\left( 
\frac{(\mathbf{q}_1^{(b)})^\top \mathbf{k}_2^{(b)}}{\sqrt{d_b}} 
\right), \\
g_{2 \leftarrow 1}^{(b)} &= \sigma\!\left( 
\frac{(\mathbf{q}_2^{(b)})^\top \mathbf{k}_1^{(b)}}{\sqrt{d_b}} 
\right),
\end{aligned}
\end{equation}
where $\mathbf{q}_1^{(b)}, \mathbf{q}_2^{(b)}, \mathbf{k}_1^{(b)}, 
\mathbf{k}_2^{(b)} \in \mathbb{R}^{d_b}$ are the projected query and key 
vectors in the $b$-th head, and $\sigma(\cdot)$ denotes the sigmoid function. 
The resulting gates $g_{1 \leftarrow 2}^{(b)}, g_{2 \leftarrow 1}^{(b)} \in (0,1)$ 
adaptively control how much information should be transferred from one 
modality to the other.

\paragraph{Gated fusion}
The gates are applied to the corresponding cross-modal value vectors to obtain modulated features:
\begin{equation}
\mathbf{a}_{1}^{(b)} = g_{1 \leftarrow 2}^{(b)} \, \mathbf{v}_2^{(b)}, \qquad
\mathbf{a}_{2}^{(b)} = g_{2 \leftarrow 1}^{(b)} \, \mathbf{v}_1^{(b)}.
\end{equation}
Finally, the outputs from all $B$ heads are concatenated within each direction and averaged across the two directions to produce the fused representation:
\begin{equation}
\mathbf{z} = \frac{
[\, \mathbf{a}_{1}^{(1)} \Vert \cdots \Vert \mathbf{a}_{1}^{(B)}\, ] +
[\, \mathbf{a}_{2}^{(1)} \Vert \cdots \Vert \mathbf{a}_{2}^{(B)}\, ]
}{2}.
\end{equation}
This bidirectional gating and symmetric aggregation ensure that both modalities contribute equitably while preserving multi-granular cross-modal interactions.

\subsection{Inter-Session Intent Transition Module}
User intent during mobile app usage is dynamic and continuously evolves with temporal context, spatial context, and recent interaction behavior. Therefore, effectively characterizing intent transitions within the current session is crucial for next app prediction. The Inter-Session Intent Transition Module models such dynamic intent by integrating the comprehensive session representation with short-term immediate intent signals to capture the user's current decision-making tendency. The comprehensive session representation encodes both multi-hop structural and contextual information and is processed by the Transformer encoder to obtain a session-level representation. The decoder then takes the immediate intent derived from recent app interactions as its query and attends to the encoder output, producing a dynamic intent representation for next app prediction.

 First, we update the comprehensive embedding by incorporating the temporal and spatial context embeddings obtained from the CMGF module, as illustrated below: 
\begin{equation}
\mathbf{{h}} = \text{CMGF}(\mathbf{g}, \text{CMGF}(\mathbf{e}^{(\tau)}_{T},\mathbf{e}^{(\rho)}_{R})).
\end{equation}
CMGF is cross-modal gated fusion module, $\mathbf{{h}} \in \mathbb{R}^{d}$. Finally, the comprehensive embedding $\mathbf{{h}} $ is transformed by an $L$-layer Transformer Encoder to obtain a context-aware session representation. The update for layer $\ell$ is
\begin{equation}
\begin{split}
\hat{\mathbf{h}}^{(0)} &= {\mathbf{{h}}}, \\
\tilde{\mathbf{h}}^{(\ell)} &= \text{LayerNorm}(\hat{\mathbf{h}}^{(\ell)} + \text{MHSA}(\hat{\mathbf{h}}^{(\ell)})), \\
\hat{\mathbf{h}}^{(\ell+1)} &= \text{LayerNorm}(\tilde{\mathbf{h}}^{(\ell)} + \text{FFN}(\tilde{\mathbf{h}}^{(\ell)})),
\end{split}
\end{equation}
where MHSA denotes multi-head self-attention, and FFN is a feed-forward network, $ \hat{\mathbf{h}}^{(L)}  $ is the final encoded session representation.

To capture the evolution of user preferences, we employ an $L$-layer Transformer decoder to model intent transitions. Instead of using a static representation, we take the short-term immediate intent $\mathbf{s}$ as the initial query state. The decoding process at layer $\ell$ is formulated as follows:
\begin{equation}
\begin{split}
\hat{\mathbf{s}}^{(0)} &= {\mathbf{s}}, \\
\tilde{\mathbf{s}_{1}}^{(\ell)} &= \text{LayerNorm}(\hat{\mathbf{s}}^{(\ell)} + \text{MHSA}(\hat{\mathbf{s}}^{(\ell)})), \\
\tilde{\mathbf{s}_{2}}^{(\ell)} &= \text{LayerNorm}(\tilde{\mathbf{s}_{1}}^{(\ell)} + \text{MHA}(\tilde{\mathbf{s}_{1}}^{(\ell)}, \hat{\mathbf{h}}^{(L)})), \\
\hat{\mathbf{s}}^{(\ell+1)} &= \text{LayerNorm}(\tilde{\mathbf{s}_{2}}^{(\ell)} + \text{FFN}(\tilde{\mathbf{s}_{2}}^{(\ell)})),
\end{split}
\end{equation}
here, MHA denotes multi-head attention. At layer $\ell$, the query is computed from $\tilde{\mathbf{s}}_1^{(\ell)}$, whereas the keys and values are obtained from the encoder's session representation $\hat{\mathbf{h}}^{(L)}$. The final output of the $L$-th layer is the representation of session intent transition
$\mathbf{u} = \hat{\mathbf{s}}^{(L)} $.

The final representation $\mathbf{u}$ produced by the decoder integrates 
(1) the structural representation aggregated from multi-hop session graphs, 
(2) the short-term immediate intent derived from the most recent apps, and 
(3) contextual information such as temporal and spatial context embeddings. $\mathbf{u}$ captures both short-term and long-term behavioral signals as well as external contextual cues, enabling accurate modeling of user preference for the next app.
The preference score for app $a_i$ is computed through the dot-product interaction: 
\begin{equation}
    \begin{split}
\hat{y}_i &= \mathbf{u}^\top \mathbf{e}_{A}(a_i), \\
\hat{y} &= [\hat{y}_1, \hat{y}_2, \hat{y}_3, \ldots, {\hat{y}}_{|\mathcal{A}|}], \\
\mathbf{P} &= \operatorname{softmax}(\hat{y}).
    \end{split}
\end{equation}

We obtain the probability $\mathbf{P}= \{\mathbf{P}_1, \mathbf{P}_2, \ldots, \mathbf{P}_{|\mathcal{A}|}\}$ of each app by applying the softmax function. The model parameters are learned during training by optimizing the cross-entropy loss: 
\begin{equation}
   \mathcal{L} = -\sum_{(S,\tau,\rho,a_i) \in \mathcal{T}} y_i \cdot \log {\mathbf{P}}_{i},
\end{equation}
where $\mathcal{T}$ denotes the training data set. $ (S,\tau,\rho,a_i) $ indicates that the next app $a_i$ to be launched after the session $S$ (with the temporal context $\tau$ and the spatial context $\rho$), $y_i$ is the ground-truth of the next app.

\section{EXPERIMENTS}
\subsection{Datasets} 
To evaluate the performance of MISApp, we test MISApp on two real-world datasets: Tsinghua App Usage~\cite{yu2018smartphone} and LSapp~\cite{aliannejadi2021context}, which are widely used for mobile app usage behavior modeling and prediction. The Tsinghua App Usage dataset additionally provides location information, which is used by MISApp as spatial context. The key dataset statistics for these two datasets are summarized in Table~\ref{tab: dataset_comparison}.
\begin{table}[htbp]
    \centering
    \caption{Comparison of Tsinghua App Usage and LSapp Datasets}
    \label{tab: dataset_comparison}
    \resizebox{\columnwidth}{!}{ 
    \begin{tabular}{lccccc}
        \toprule
        \textbf{Dataset} & \textbf{Collection Date} & \textbf{Users} & \textbf{Apps} & \textbf{Records} &  \textbf{Attributes} \\
        \midrule
        \multirow{3}{*}{Tsinghua App Usage} & \multirow{3}{*}{19--26 April 2016} & \multirow{3}{*}{1000} & \multirow{3}{*}{2000} & \multirow{3}{*}{4, 171, 950} & User ID, Timestamp, \\ 
        & & & & & Base Station ID, App ID, \\ 
        & & & & & App Category ID, Category Label\\
        \midrule
        \multirow{3}{*}{LSapp} & \multirow{3}{*}{\makecell{9 September 2017- \\ 17 May 2018}} & \multirow{3}{*}{292} & \multirow{3}{*}{87} & \multirow{3}{*}{3, 658, 590} & User ID, Timestamp, \\ 
        & & & & & Session ID, App ID, \\ 
        & & & & & Interaction Type\\
        \bottomrule
    \end{tabular}}
\end{table}

\subsection{Experimental Settings}
During preprocessing, consecutive records of the same app by the same user with an interval of less than 5 minutes were merged to reduce fragmentation. Users with fewer than 50 total app usage events were removed to ensure sufficient data density. Based on the 300 seconds threshold, continuous events were grouped into sessions. Sessions longer than 5000 records were treated as noise and discarded. These steps produce a clean, well-structured dataset suitable for subsequent modeling. 
 For both datasets, we adopt the same cleaning procedures and splitting protocol as used in MAPLE~\cite{khaokaew2024maple}.
\begin{itemize}
\item
Standard Setting: 
For each user, data is divided chronologically into 70\% for training, 10\% for validation, and 20\% for testing. This configuration supports sufficient model learning, parameter tuning, and performance evaluation in a time-aware manner.
\item 
Cold-Start Setting: 
To assess MISApp’s ability to generalize to unseen users, we adopt a user-level split, allocating 90\% of users for training and 10\% for testing. To ensure a fair comparison, all splits share the same app set, and no new apps are introduced in the test phase.
\end{itemize}

To evaluate the effectiveness of the next app prediction, we employ two widely used ranking-based metrics: Accuracy@k (ACC@k) and Mean Reciprocal Rank (MRR@k), both reported as percentages (\%). These metrics assess how well the model ranks the true next app among all candidates.
\noindent
ACC@k measures the proportion of test instances where the ground-truth next app appears within the top-k predictions. It is defined as
\begin{equation}
    \text{ACC@k} = 
    \frac{1}{N_{test}} \sum_{i=1}^{N_{test}} \mathbb{I}\left( \text {rank}_i \leq \text{k} \right), 
\end{equation}
where $ N_{test}$ denotes the number of test instances and $\text {rank}_i$ is the rank of the i-th correct prediction.
\noindent
MRR@k evaluates how highly the model ranks the correct next app in the predicted list by computing the reciprocal rank: 
\begin{equation}
\text{MRR@k} =
\frac{1}{N_{\mathrm{test}}}
\sum_{i=1}^{N_{\mathrm{test}}}
\frac{\mathbb{I}(\mathrm{rank}_i \le \text{k})}
{\mathrm{rank}_i}.
\end{equation}

\noindent

We compare MISApp against a comprehensive set of baselines spanning three categories: app usage prediction models, graph-based models, and time-series forecasting architectures.

\begin{itemize}

\item \textbf{App Usage Prediction Models}: 
MFU/MRU~\cite{shin2012understanding} rely on simple frequency- and recency-based heuristics. 
AppUsage2Vec~\cite{zhao2019appusage2vec}, DeepApp~\cite{xia2020deepapp}, DeepPattern~\cite{suleiman2021DeepPatterns}, Appformer~\cite{sun2025appformer} and NeuSA~\cite{aliannejadi2021context} employ sequential learning to model contextual app transitions. 
CoSEM~\cite{khaokaew2021cosem} and MAPLE~\cite{khaokaew2024maple} incorporate contextual and semantic embeddings, with MAPLE further leveraging LLM-based representations to enhance prediction.

\item \textbf{Graph-based Prediction Models}: 
SA-GCN~\cite{yu2020semantic} and Hypergraph-based approaches~\cite{huang2025predicting} model relational dependencies among apps through graph or hypergraph structures, enabling higher-order connectivity reasoning in app usage patterns.

\item \textbf{Time-Series Forecasting Models}: 
Transformer variants—including Transformer~\cite{vaswani2017attention}, Reformer~\cite{kitaev2020reformer}, FEDformer~\cite{zhou2022fedformer}, TimesNet~\cite{wu2022timesnet}—provide strong baselines designed primarily for long-sequence temporal modeling. 
We additionally include lightweight yet competitive non-Transformer forecasting baselines such as DLinear~\cite{zeng2023transformers}, which adopt linear decomposition strategies for efficient trend–season separation.

\end{itemize}
\noindent 

For the Tsinghua App Usage dataset, the available location-related information is used when required by these models. 
For the LSapp dataset, since location information is unavailable, the location-related branch or input feature is removed, and these models are evaluated using only the available app usage and temporal information. 
On the Tsinghua App Usage dataset, the Hypergraph-based model adopts the same graph-clustering-based spatial embedding strategy as used in our model for a fair comparison.

MISApp is configured with an embedding dimension of 64 for app embeddings and temporal context embeddings. 
For the Tsinghua App Usage dataset, where base station and POI information is available, the spatial context embedding is also set to 64 dimensions. 
For the LSapp dataset, since spatial context information is unavailable, the spatial branch is removed. 
Accordingly, in the Inter-Session Intent Transition module, the comprehensive embedding is obtained by applying CMGF only to the multi-hop structural representation $\mathbf{g}$ and the temporal context embedding $\mathbf{e}^{(\tau)}_{T}$:
\begin{equation}
\mathbf{h} = \mathrm{CMGF}\left(\mathbf{g}, \mathbf{e}^{(\tau)}_{T}\right).
\end{equation}

The immediate intent window size is set to $K=3$, and the session length is fixed to $T=8$. 
The Transformer encoder consists of 2 layers with 4 attention heads, and the decoder has the same architecture. 
The Cross-Modal Gated Fusion module employs 4 heads. 
The LightGCN module is set to $L_g=2$ propagation layers. 
Training is performed with a batch size of 512, a learning rate of 0.001, and a dropout rate of 0.10. 
The model is optimized using the Adam optimizer and trained for 50 epochs. All reproducible models are trained on a single NVIDIA GeForce RTX 4090 GPU and evaluated over five independent runs, with the mean performance and standard deviation reported. Sensitivity analyses of the immediate intent window size, embedding dimension, maximum session window size, and number of Transformer layers are reported in Section~S4 of the supplementary material.
\subsection{Performance Comparison}
\paragraph{Results on Standard Setting} As shown in Table~\ref{tab: standard_split_results}, MISApp consistently outperforms existing models on both the Tsinghua App Usage and LSapp datasets under the standard evaluation protocol, achieving superior performance across all metrics and demonstrating the effectiveness of its design for the next app prediction in standard settings.
\begin{table*}[!t]
\centering
\caption{Comparison results under the standard data split. 
Each experiment is repeated five times, and the results are reported as mean$_{\pm{std}}$. 
The best results are highlighted in bold, and the second-best results are underlined.}
\resizebox{1.0\linewidth}{!}{
\begin{tabular}{lcccc c|ccccc}
\toprule
{\textbf{Datasets}} 
& \multicolumn{5}{c}{\textbf{Tsinghua App Usage}} 
& \multicolumn{5}{c}{\textbf{LSapp}} \\
\midrule
\multirow{1}{*}{\textbf{Models}} 

& \textbf{ACC@1} & \textbf{ACC@3} & \textbf{ACC@5} & \textbf{MRR@3} & \textbf{MRR@5}
& \textbf{ACC@1} & \textbf{ACC@3} & \textbf{ACC@5} & \textbf{MRR@3} & \textbf{MRR@5} \\
\midrule

MFU
& 18.41$_{\pm0.000}$ & 38.17$_{\pm0.000}$ & 48.81$_{\pm0.000}$ &  27.00$_{\pm0.000}$ & 29.42$_{\pm0.000}$ 
& 30.24$_{\pm0.000}$ & 62.39$_{\pm0.000}$ & 78.35$_{\pm0.000}$ & 44.18$_{\pm0.000}$ &  47.85$_{\pm0.000}$\\

MRU
& 7.77$_{\pm0.000}$&  20.06$_{\pm0.000}$ & 26.15$_{\pm0.000}$ & 13.23$_{\pm0.000}$ & 14.60$_{\pm0.000}$ 
& 20.01$_{\pm0.000}$ & 39.22$_{\pm0.000}$ & 51.70$_{\pm0.000}$ & 27.91$_{\pm0.000}$ & 30.76$_{\pm0.000}$\\

AppUsage2Vec
& 39.32$_{\pm0.322}$ & 64.45$_{\pm0.155}$ & 76.23$_{\pm0.120}$ & 50.26$_{\pm0.222}$ & 52.95$_{\pm0.204}$
& 81.95$_{\pm0.126}$ & 91.82$_{\pm0.089}$ & 94.65$_{\pm0.041}$ & 85.43$_{\pm0.084}$ & 86.33$_{\pm0.076}$\\

Transformer
& 48.24$_{\pm0.018}$ & 67.28$_{\pm0.111}$ & 73.49$_{\pm0.032}$ &56.70$_{\pm0.016}$ & 58.14$_{\pm0.026}$
& 79.88$_{\pm0.167}$ & 89.95$_{\pm0.172}$ & 92.08$_{\pm0.194}$ & 84.58$_{\pm0.163}$ & 84.69$_{\pm0.168}$ \\

Reformer
& 48.07$_{\pm0.048}$ & 66.66$_{\pm0.080}$ & 73.07$_{\pm0.094}$ & 56.35$_{\pm0.055}$ & 57.82 $_{\pm0.059}$
& 80.18$_{\pm0.139}$ & 88.59$_{\pm0.198}$ & 91.83$_{\pm0.185}$ & 84.54$_{\pm0.154}$& 84.86$_{\pm0.151}$ \\

TimesNet
& 51.83$_{\pm0.070}$ & 72.46$_{\pm0.046}$ & 79.24$_{\pm0.044}$ & 60.50$_{\pm0.043}$ & 62.06$_{\pm0.044}$ 
& \underline{83.88}$_{\pm0.066}$  & \underline{92.33}$_{\pm0.079}$ & 94.35$_{\pm0.124}$ & 87.32$_{\pm0.070}$ & 88.06$_{\pm0.071}$ \\

FEDformer
& \underline{52.89}$_{\pm0.018}$ & 73.60$_{\pm0.159}$ & 80.39$_{\pm0.204}$ & \underline{62.58}$_{\pm0.127}$ & {63.14}$_{\pm0.138}$
& {83.86}$_{\pm0.034}$ & {92.30}$_{\pm0.168}$ & \underline{94.44}$_{\pm0.149}$ & \underline{87.71} $_{\pm0.093}$& \underline{88.20}$_{\pm0.087}$ \\

DLinear
& 10.43$_{\pm0.221}$ & 19.99$_{\pm0.086}$ & 26.98$_{\pm0.071}$ & 14.98$_{\pm0.253}$ & 15.82$_{\pm0.059}$ 
& 24.63$_{\pm0.067}$ & 51.57$_{\pm0.048}$ & 67.13$_{\pm0.020}$ & 35.40$_{\pm0.040}$ & 39.91$_{\pm0.043}$\\

DeepApp
& 28.14$_{\pm0.119}$ & 64.08$_{\pm0.158}$  & 71.85$_{\pm0.155}$  & 42.51$_{\pm0.130}$  & 44.29$_{\pm0.126}$ 
& 75.87$_{\pm0.507}$ & 87.04$_{\pm0.114}$& 90.76$_{\pm0.088}$ & 80.76$_{\pm0.338}$ & 81.61$_{\pm0.316}$ \\

DeepPattern
& 38.91$_{\pm0.359}$ & 56.04$_{\pm0.391}$  & 64.03$_{\pm0.365}$  & 46.42$_{\pm0.374}$  & 48.25$_{\pm0.330}$ 
&  64.20$_{\pm1.667}$ & 83.44$_{\pm0.204}$ & 88.84$_{\pm0.113}$ & 72.65$_{\pm0.999}$ & 73.90$_{\pm0.957}$ \\

SA-GCN
& 46.55$_{\pm0.224}$ & 67.65$_{\pm0.362}$ & 75.22$_{\pm0.324}$ & 55.94$_{\pm0.267}$ & 57.68$_{\pm0.264}$
& 79.30$_{\pm0.979}$ & 85.13$_{\pm0.435}$ & 87.97$_{\pm0.244}$ & 81.90$_{\pm0.722}$ & 82.55$_{\pm0.669}$ \\ 

NeuSA
& 46.40 & 65.62 & 72.86 & 54.92 & 56.58
& 68.32 & 82.53 & 88.30 & 74.61 & 75.93 \\

CoSEM
& 41.63 & 66.82 & 74.99 & 52.82 & 54.69
& 49.90 & 74.66 & 81.49 & 60.83 & 62.42 \\

MAPLE
& 51.91 & \underline{73.85} & \underline{81.15} & 61.69 & \underline{63.38}
& 71.57 & 86.49 & 91.50 & 78.21 & 79.36 \\

Hypergraph Model
& 47.52$_{\pm0.117}$ & 69.82$_{\pm0.290}$ & 77.47$_{\pm0.220}$ & 57.72$_{\pm0.257}$ & 59.26$_{\pm0.237}$
& 82.87$_{\pm0.355}$ & 91.27$_{\pm0.238}$ & 93.46$_{\pm0.322}$ & 86.61$_{\pm0.122}$ & 87.11$_{\pm0.134}$   \\

Appformer
& 27.30$_{\pm1.532}$ & 48.10$_{\pm1.762}$ & 57.40$_{\pm1.769}$ & 36.70$_{\pm0.426}$ & 38.74$_{\pm0.504}$
& 81.67$_{\pm0.131}$ & 90.80$_{\pm0.549}$ & 93.20$_{\pm0.637}$ & 85.80$_{\pm0.320}$ & 86.36$_{\pm0.336}$\\

\midrule
\textbf{MISApp}
&  \textbf{54.52$_{\pm0.175}$} & \textbf{76.33$_{\pm0.271}$} & \textbf{82.40$_{\pm0.240}$} & \textbf{64.26$_{\pm0.214}$} & \textbf{65.89$_{\pm0.209}$}  & 
\textbf{84.00$_{\pm0.038}$} & \textbf{92.44$_{\pm0.167}$} & \textbf{94.87$_{\pm0.142}$} & \textbf{87.93$_{\pm0.095}$} & \textbf{88.63$_{\pm0.090}$}   \\

\bottomrule
\end{tabular}
}
\begin{flushleft}
\scriptsize
\textit{Note.} The results of NeuSA, CoSEM, and MAPLE are taken from the MAPLE paper, since their complete reproducible code and detailed configurations are not publicly available. Standard deviations are therefore not reported for these methods. 
\end{flushleft}
\label{tab: standard_split_results}
\end{table*}

\begin{table*}[!t]
\centering
\caption{Comparison results under the cold-start split. Each experiment is repeated five times, and the results are reported as mean$_{\pm{std}}$. 
The best results are highlighted in bold, and the second-best results are underlined.}
\resizebox{1.0\linewidth}{!}{
\begin{tabular}{lcccc c|ccccc}
\toprule
{\textbf{Datasets}} 
& \multicolumn{5}{c}{\textbf{Tsinghua App Usage}} 
& \multicolumn{5}{c}{\textbf{LSapp}} \\
\midrule
\multirow{1}{*}{\textbf{Models}} 

& \textbf{ACC@1} & \textbf{ACC@3} & \textbf{ACC@5} & \textbf{MRR@3} & \textbf{MRR@5}
& \textbf{ACC@1} & \textbf{ACC@3} & \textbf{ACC@5} & \textbf{MRR@3} & \textbf{MRR@5} \\
\midrule

MFU
& 18.40$_{\pm0.000}$ & 39.62$_{\pm0.000}$ & 51.43$_{\pm0.000}$ & 27.59$_{\pm0.000}$ & 30.29$_{\pm0.000}$ 
& 28.00$_{\pm0.000}$ & 64.33$_{\pm0.000}$ & 82.33$_{\pm0.000}$ & 43.73$_{\pm0.000}$ & 47.86$_{\pm0.000}$ \\

MRU
&  0.00$_{\pm0.000}$ & 56.49$_{\pm0.000}$ & 66.61$_{\pm0.000}$ & 26.37$_{\pm0.000}$ & 28.72$_{\pm0.000}$
& 0.00$_{\pm0.000}$ & 85.49$_{\pm0.000}$ & 88.89$_{\pm0.000}$ & 42.08$_{\pm0.000}$  & 42.87$_{\pm0.000}$\\

AppUsage2Vec
& 24.22$_{\pm2.488}$ & 41.19$_{\pm3.728}$ & 51.13$_{\pm4.523}$ & 31.53$_{\pm3.000}$ & 33.79$_{\pm3.170}$
& 53.10$_{\pm2.082}$ & 61.15$_{\pm0.933}$ & 63.87$_{\pm0.642}$ & 56.73$_{\pm1.483}$ & 57.56$_{\pm1.382}$ \\

Transformer
& 48.77$_{\pm0.194}$ & 68.07$_{\pm0.054}$ & 74.09$_{\pm0.025}$ & 57.44$_{\pm0.133}$ & 58.82$_{\pm0.121}$
& 80.16$_{\pm0.054}$ & 89.52$_{\pm0.068}$ & 91.74$_{\pm0.075}$ & 84.47$_{\pm0.030}$ & 84.98$_{\pm0.034}$ \\

Reformer
& 48.59$_{\pm0.093}$ & 67.67$_{\pm0.178}$ & 73.71$_{\pm0.190}$ & 57.17$_{\pm0.120}$ & 58.56$_{\pm0.122}$
& 80.42$_{\pm0.135}$ & 89.01$_{\pm0.174}$ & 91.19$_{\pm0.123}$ & 84.34$_{\pm0.094}$& 84.84$_{\pm0.091}$ \\

TimesNet
& 51.53$_{\pm0.176}$ & 72.88$_{\pm0.181}$ & 78.20$_{\pm0.153}$ & 60.44$_{\pm0.166}$ & 62.89$_{\pm0.165}$
& 81.26$_{\pm0.041}$ & {90.60}$_{\pm0.235}$ & {92.74}$_{\pm0.249}$ & \underline{85.54}$_{\pm0.120}$ & {86.03}$_{\pm0.105}$ \\

FEDformer
& 52.19$_{\pm0.108}$ & 73.11$_{\pm0.167}$ & 80.40$_{\pm0.110}$ & \underline{62.59}$_{\pm0.096}$ & \underline{63.83}$_{\pm0.081}$  
& \underline{81.39}$_{\pm0.034}$ & \underline{90.80}$_{\pm0.366}$ & \underline{92.99}$_{\pm0.425}$ & 85.48$_{\pm0.150}$ & \underline{86.08}$_{\pm0.158}$\\

DLinear
& 7.83$_{\pm0.174}$ & 17.36$_{\pm0.198}$ & 23.77$_{\pm0.269}$ & 11.83$_{\pm0.174}$ & 13.50$_{\pm0.172}$   
& 30.05$_{\pm0.090}$ & 52.26$_{\pm0.243}$ & 64.64$_{\pm0.301}$ & 40.27$_{\pm0.117}$ & 43.81$_{\pm0.126}$\\

DeepApp
& 29.46$_{\pm0.135}$ & 65.35$_{\pm0.191}$  & 72.37$_{\pm0.150}$  & 54.86$_{\pm0.119}$  & 56.47$_{\pm0.075}$
& 75.72$_{\pm0.503}$ & 86.59$_{\pm0.207}$ & 89.80$_{\pm0.246}$ & 80.51$_{\pm0.350}$ & 81.25$_{\pm0.277}$ \\

DeepPattern
& 35.14$_{\pm0.119}$ & 64.08$_{\pm0.158}$  & 71.85$_{\pm0.155}$  & 53.51$_{\pm0.130}$  & 55.29$_{\pm0.126}$ 
&  64.25$_{\pm1.194}$ & 81.56$_{\pm0.495}$ & 86.28$_{\pm0.285}$ & 71.78$_{\pm0.772}$ & 72.87$_{\pm0.733}$ \\


SA-GCN
& 45.80$_{\pm0.176}$ & 65.68$_{\pm0.170}$ & 72.88$_{\pm0.307}$ & 54.67$_{\pm0.158}$ & 56.32$_{\pm0.183}$
& 76.72$_{\pm1.211}$ & 83.02$_{\pm0.933}$ & 86.03$_{\pm0.613}$ & 79.52$_{\pm1.082}$ & 80.21$_{\pm1.006}$ \\


NeuSA
& 44.33 & 61.69 & 68.12 & 52.06 & 53.53
& 68.74 & 77.70 & 81.35 & 72.72 & 73.55 \\

CoSEM
& 31.11 & 55.97 & 65.25 & 42.04 & 44.16
& 45.23 & 72.43 & 81.04 & 57.18 & 59.18 \\

MAPLE
& \underline{52.28} & \underline{74.17} & \underline{81.28} & {62.06} & {63.60}
& 76.44 & 88.48 & 92.47 & 81.81 & 82.72 \\

Hypergraph Model
& 48.41$_{\pm0.125}$ & 68.55$_{\pm0.346}$ & 75.14$_{\pm0.428}$ & 57.57$_{\pm0.343}$ & 58.09$_{\pm0.363}$ 
& 80.83$_{\pm0.207}$ & 89.59$_{\pm0.180}$ & 91.99$_{\pm0.317}$ & 84.25$_{\pm0.174}$ & 84.81$_{\pm0.201}$\\

Appformer
&  28.07$_{\pm0.870}$ & 48.80$_{\pm0.865}$ &  59.05$_{\pm0.914}$& 36.44$_{\pm0.220}$ & 38.98$_{\pm0.144}$
& 80.92$_{\pm0.129}$ & 89.99$_{\pm0.428}$ & 92.14$_{\pm0.569}$ & 84.86$_{\pm0.368}$ & 85.39$_{\pm0.383}$\\

\midrule
\textbf{MISApp}
& \textbf{54.64$_{\pm0.281}$} 
& \textbf{76.26$_{\pm0.401}$} 
& \textbf{81.90$_{\pm0.383}$} 
& \textbf{64.42$_{\pm0.341}$} 
& \textbf{64.91$_{\pm0.334}$}
& \textbf{81.50$_{\pm0.039}$} 
& \textbf{91.80$_{\pm0.168}$} 
& \textbf{93.75$_{\pm0.125}$} 
& \textbf{86.55$_{\pm0.078}$} 
& \textbf{87.49$_{\pm0.051}$} \\

\bottomrule
\end{tabular}
}
\label{tab: coldstart_split_results}
\end{table*}

Traditional app usage models such as AppUsage2Vec, DeepApp, DeepPattern, NeuSA, and CoSEM improve over heuristics, yet their reliance on static embeddings or isolated sequence encoders limits their ability to capture rich contextual cues. Transformer-based models (Transformer, Reformer, FEDformer, and Appformer) enhance sequential modeling, but may still overlook heterogeneous factors such as temporal shifts and cross-app relational patterns.

The Hypergraph Model achieves strong performance by capturing higher-order relations, but conventional hyperedges lack internal ordering and thus cannot explicitly preserve sequential transition structures. Similarly, stacking multiple GCN layers may introduce over-smoothing, whereas our multi-hop design explicitly separates different hop-level dependencies without deep propagation. MAPLE also performs well by leveraging large-scale pretrained representations, but at the cost of greater model complexity. In contrast, MISApp integrates multi-hop session graphs, lightweight LightGCN-based structural modeling, and dynamic intent modeling into a complementary and efficient framework. Controlled comparisons with conventional graph, merged-graph, and hypergraph variants in Section~S2 of the supplementary material further validate the benefit of explicitly separating and adaptively fusing different hop ranges.

\begin{table*}[ht] 
\centering
\caption{Ablation Study.}
\resizebox{1.0\linewidth}{!}{
\begin{tabular}{lcccc c|ccccc}
\toprule
{\textbf{Datasets}} 
& \multicolumn{5}{c}{\textbf{Tsinghua App Usage}} 
& \multicolumn{5}{c}{\textbf{LSapp}} \\
\midrule
\multirow{1}{*}{\textbf{Models}} 

& \textbf{ACC@1} & \textbf{ACC@3} & \textbf{ACC@5} & \textbf{MRR@3} & \textbf{MRR@5}
& \textbf{ACC@1} & \textbf{ACC@3} & \textbf{ACC@5} & \textbf{MRR@3} & \textbf{MRR@5} \\
\midrule

w/o Multi-hop Graph
&51.56$_{\pm0.062}$ & 71.92$_{\pm0.045}$ & 80.91$_{\pm0.019}$ & 62.02$_{\pm0.048}$ & 63.86$_{\pm0.044}$ 
            & 51.82$_{\pm0.200}$ & 90.90$_{\pm0.019}$ & 94.13$_{\pm0.043}$ & 70.47$_{\pm0.100}$ & 71.24$_{\pm0.095}$ \\

w/o Temporal Context
& 53.23$_{\pm0.117}$ & 73.26$_{\pm0.150}$ & 80.51$_{\pm0.140}$ & 62.19$_{\pm0.130}$ & 63.82$_{\pm0.129}$ 
& 83.02$_{\pm0.052}$ & 91.27$_{\pm0.203}$ & 93.48$_{\pm0.258}$ & 87.26$_{\pm0.115}$ & 87.57$_{\pm0.126}$ \\

w/o Spatial Context
& 53.56$_{\pm0.095}$ & 74.39$_{\pm0.121}$ & 81.48$_{\pm0.131}$ & 62.31$_{\pm0.107}$ & 63.94$_{\pm0.108}$
& -- & -- & -- & -- & --\\

w/o Decoder
& 43.52$_{\pm1.072}$  & 68.46$_{\pm1.481}$  & 78.63$_{\pm0.365}$  & 54.78$_{\pm0.825}$  & 56.97$_{\pm0.781}$  
& 77.02$_{\pm0.052}$ & 91.27$_{\pm0.203}$ & 93.48$_{\pm0.258}$ & 83.76$_{\pm0.115}$ & 84.27$_{\pm0.126}$ \\

\midrule
\textbf{MISApp}
& \textbf{54.52$_{\pm0.175}$} & \textbf{76.33$_{\pm0.271}$} & \textbf{82.40$_{\pm0.240}$} & \textbf{64.26$_{\pm0.214}$} & \textbf{65.89$_{\pm0.209}$}  & 
\textbf{84.00$_{\pm0.038}$} & \textbf{92.44$_{\pm0.167}$} & \textbf{94.87$_{\pm0.142}$} & \textbf{87.93$_{\pm0.095}$} & \textbf{88.63$_{\pm0.090}$}  \\

\bottomrule
\end{tabular}
}
\label{tab: Ablation Study}
\end{table*}

\paragraph{Results on Cold-Start Setting}
Under the cold-start evaluation protocol, MISApp also achieves the best performance across all metrics on both datasets (Table~\ref{tab: coldstart_split_results}), with its benefits particularly pronounced when user history is limited. Methods that rely on personalized embeddings or long-term behavioral patterns—such as AppUsage2Vec and other profile-driven baselines—tend to show reduced performance in this setting, as cold-start users lack sufficient historical data to establish stable preference priors. Transformer-style sequence models also show limited effectiveness: by treating app usage as uniformly evolving temporal signals, they struggle to extract reliable intent from extremely short and noisy interaction sequences.

MISApp does not rely on user-specific profiles or long-term interaction histories. Its prediction is instead derived from the multi-hop structure of the current session together with immediate contextual signals. This formulation is particularly useful for unseen users, for whom reliable long-term preference estimates are unavailable. As a result, MISApp consistently demonstrates superior adaptability and resilience in cold-start scenarios.

\subsection{Ablation Study}
To assess the contribution of each core component, we conduct an ablation study by progressively removing key modules from MISApp. Four variants are evaluated, and the results on the standard split are shown in Table~\ref{tab: Ablation Study}.

\paragraph{Without multi-hop graph structure (w/o Multi-hop Gra
ph)} This variant removes the multi-hop structural components and retains only the traditional 1-Hop session graph. The performance degradation demonstrates that higher-order structural dependencies contribute meaningful relational signals rather than introducing noise. By explicitly modeling multi-hop transitions, MISApp preserves distinct dependency ranges within the session, which are essential for capturing semantic progression and dynamic intent shifts during app transitions.

\paragraph{Without temporal context (w/o Temporal Context)}
Removing temporal context embedding deprives MISApp of fine-grained temporal cues that reflect the dynamics of app interactions within a session. The moderate performance drop suggests that while temporal information provides useful contextual signals, the model’s core predictive power primarily stems from structural and intent-aware representations.

\paragraph{Without spatial context (w/o Spatial Context)}
Removing spatial context also leads to a clear performance degradation on
the Tsinghua App Usage dataset, confirming that spatial information provides
useful complementary cues for modeling activity-dependent app usage.
Together with the temporal-context ablation, these results show that both
contextual modalities contribute to the final prediction performance.

\paragraph{Without dynamic intent decoder (w/o Decoder)}
We disable MISApp’s dynamic intent decoder, relying only on the encoder’s static session representation. This variant suffers the largest performance drop, highlighting that modeling evolving user intent is essential for capturing evolving session intent.

\subsection{Modal Fusion Study}
To evaluate the effectiveness of the Cross-Modal Gated Fusion module, we compare it against several alternative fusion strategies: Sum Fusion, Mean Fusion, CNN Fusion, MLP Fusion and Gated Fusion. The results on the Tsinghua App Usage dataset under the standard split are summarized in Table~\ref{tab:fusion_methods}.

\begin{table}[http]
\centering
\scriptsize
\setlength{\tabcolsep}{2pt}
\renewcommand{\arraystretch}{1.0}
\caption{Performance comparison of different modal fusion strategies. The best results are highlighted in bold, and the second-best results are underlined.}
\label{tab:fusion_methods}
\renewcommand{\arraystretch}{1.1} 
\begin{tabular}{l|ccccc} %
\hline
\addlinespace[4pt] 
Methods & ACC@1 & ACC@3 & ACC@5 & MRR@3 & MRR@5 \\
\addlinespace[2pt] 
\hline
Sum    & 53.27$_{\pm0.036}$ & 74.31$_{\pm0.040}$ & 80.93$_{\pm0.030}$ & 62.63$_{\pm0.031}$ & 64.05$_{\pm0.028}$ \\
Mean   & 53.42$_{\pm0.041}$ & 74.69$_{\pm0.062}$ & 80.70$_{\pm0.054}$ & 62.66$_{\pm0.041}$ & 64.20$_{\pm0.039}$ \\
CNN    & \underline{53.86}$_{\pm0.061}$ & 75.66$_{\pm0.111}$ & 82.22$_{\pm0.049}$ & 63.62$_{\pm0.069}$ & 64.96$_{\pm0.056}$ \\
MLP    & 53.26$_{\pm0.042}$ & 73.95$_{\pm0.054}$ & 80.51$_{\pm0.034}$ & 62.45$_{\pm0.034}$ & 63.99$_{\pm0.031}$ \\
Gated  & 53.82$_{\pm0.041}$ & \underline{75.69}$_{\pm0.062}$ & \textbf{82.40}$_{\pm0.054}$ & \underline{63.66}$_{\pm0.041}$ & \underline{65.20}$_{\pm0.039}$ \\
\hline
CMGF   & \textbf{54.52$_{\pm0.175}$} & \textbf{76.33$_{\pm0.271}$} & \textbf{82.40$_{\pm0.240}$} & \textbf{64.26$_{\pm0.214}$} & \textbf{65.89$_{\pm0.209}$} \\
\hline
\end{tabular}
\end{table}

The experimental results show that our method achieves superior performance. Mean Fusion and Sum Fusion directly combine modality representations through element-wise averaging or addition. 
Although these strategies are simple and computationally efficient, they treat all modalities equally and ignore the varying importance of different modalities, which may lead to suboptimal representations when the contributions of modalities are imbalanced. CNN Fusion first transforms two one-dimensional modality features into a two-dimensional feature representation and then applies convolution operations to model their interactions. However, convolutional operations  focus on local patterns, which may limit their ability to capture global dependencies between modality representations. MLP Fusion concatenates modality features and learns a joint representation through a multilayer perceptron, enabling nonlinear feature interactions. 
However, it lacks an explicit mechanism to regulate modality importance and may introduce redundant information during the fusion process. 
Gated Fusion introduces a gating mechanism to dynamically control the contribution of each modality by learning adaptive weights, but the gating scores are usually computed independently and cannot explicitly capture cross-modal interactions.
In contrast, the Cross-Modal Gated Fusion explicitly models cross-modal interactions through a gated attention mechanism, enabling the model to adaptively regulate modality contributions while capturing their mutual dependencies, thereby achieving superior performance.


\begin{table*}[ht]
\centering
\caption{Comparison of practical efficiency across app prediction methods. The efficiency measurements are obtained on our experimental platform equipped
with an NVIDIA RTX 4090 GPU under the same evaluation protocol. For fair
comparison, the batch size and embedding dimension are kept consistent with
MISApp, while the remaining hyperparameters are set according to the optimal
configurations of the corresponding models.}
\label{tab:efficiency}
\small
\setlength{\tabcolsep}{4pt}
\renewcommand{\arraystretch}{0.9}

\begin{tabular}{l|cccccccc}
\toprule
Models &
\makecell{Parameters \\ (M)} &
\makecell{Model size \\ (MB)} &
\makecell{Memory \\ (MB)} &
\makecell{Training time \\ (min)} &
\makecell{Latency Mean \\ (ms)} &
\makecell{P95 Latency \\ (ms)} &
\makecell{Inference time \\ (s/sample)} &
\makecell{ACC@1} \\
\midrule

DeepApp
& 2.17 & 8.29 & 547 & 11.74
& 0.6648 & 0.7348 & 0.0005e-2 & 28.14 \\

DeepPattern
& 1.71 & 6.51 & 589 & 10.02
& 0.5054 & 0.5394 & 0.0017e-2 & 38.91 \\

Transformer
& 2.18 & 8.30 & 312 & 13.77
& 4.57 & 15.64 & 0.0009e-2 & 48.24 \\

Reformer
& 2.12 & 8.11 & 630 & 21.23
& 7.55 & 10.00 & 0.0014e-2 & 48.07 \\

TimesNet
& 11.15 & 42.55 & 324 & 34.97
& 21.62 & 31.82 & 0.0030e-2 & 51.83 \\

FEDformer
& 2.41 & 9.20 & 353 & 38.63
& 9.85 & 15.62 & 0.0017e-2 & 52.89 \\

Appformer
& 1.48 & 5.68 & 1017 & 494.16
& 286.69 & 364.39 & 0.0550e-2 & 27.30 \\

Hypergraph Model
& 1.55 & 7.95 & 1006 & 148.33
& 195.70 & 200.03 & 0.0376e-2 & 47.52 \\

MAPLE
& 60.5 & 2142.9 & -- & --
& 699.60 & 789.00 & 0.7001e-2 & 51.91 \\

\midrule

MISApp
& 1.32 & 5.08 & 435 & 29.20
& 38.17 & 43.67 & 0.0066e-2 & 54.52 \\

\bottomrule
\end{tabular}

\vspace{1pt}
\begin{minipage}{0.98\linewidth}
\footnotesize
\textit{Note.}
For MAPLE, ACC@1 is taken from the original paper, while its inference efficiency is measured using simulated inputs. Training time and peak GPU memory are omitted because MAPLE relies on a pretrained large language model rather than end-to-end training under our experimental setting.
\end{minipage}

\end{table*}

\subsection{Efficiency and Deployment Analysis}

To further evaluate the practicality of existing models for mobile application prediction, we analyze several representative methods in terms of training cost, inference efficiency, and prediction accuracy, as summarized in Table~\ref{tab:efficiency}. The compared methods include lightweight app prediction models such as DeepApp and DeepPattern, Transformer-based time-series models including Transformer, Reformer, TimesNet, and FEDformer, as well as stronger graph- or context-aware baselines such as Appformer, Hypergraph Model, and MAPLE. Specifically, we report the number of model parameters, model size, GPU memory consumption during training, training time, mean inference latency, 95th-percentile inference latency, inference time per sample, and ACC@1 on the Tsinghua App Usage dataset under the standard split.

Regarding inference efficiency, lightweight baselines such as DeepApp and DeepPattern achieve very low latency due to their relatively simple architectures. However, their prediction accuracy remains limited. Transformer-based models provide stronger predictive performance, but their computational cost varies depending on the feature extraction architecture. More advanced methods such as Appformer, Hypergraph Model, and MAPLE introduce richer contextual or structural modeling, yet they often require higher memory consumption or inference latency.

In contrast, MISApp achieves the highest ACC@1 while maintaining the smallest model size and relatively low training memory consumption. Its training time is also lower than that of several more complex baselines, such as TimesNet, FEDformer, Appformer, and Hypergraph Model. Although its inference latency is moderately higher than that of some lightweight baselines, it remains much lower than Appformer, Hypergraph Model, and MAPLE, and is still acceptable for practical online prediction scenarios.

Overall, the results show that MISApp achieves a favorable balance between predictive accuracy and computational efficiency, making it suitable for real-world deployment in intelligent mobile service prediction systems where both performance and resource efficiency are important.

\begin{table}[htbp]
\centering
\caption{Statistical Distribution of Kendall's Tau ($\tau$) Correlation.}
\label{tab:kendall_distribution}
\begin{tabular}{lccc}
\toprule
\textbf{Correlation Range} & \textbf{Sample Count} & \textbf{Percentage (\%)} \\ \midrule
$[0.9, 1.0]$  & 5 & 10\% \\
$[0.8, 0.9)$  & 30 & 60\% \\
$[0.5, 0.8)$  & 7 & 14\% \\
$< 0.5$  & 8 & 16\% \\ 
\bottomrule
\end{tabular}
\end{table}

\begin{table*}[ht]
\centering
\caption{Alignment results for the correlation between hop weights and similarity metrics across sample sequences}
\begin{tabular}{l|c|c|ccc|ccc}
\toprule
\textbf{Sequence ID} & 
\textbf{App IDs} & 
\textbf{Target App ID} & 
\textbf{$\mathbf{sim}_1$} & 
\textbf{$\mathbf{sim}_2$} & 
\textbf{$\mathbf{sim}_3$} & 
\textbf{$\mathbf{W}_{\text{hop1}}$} & 
\textbf{$\mathbf{W}_{\text{hop2}}$} & 
\textbf{$\mathbf{W}_{\text{hop3}}$} \\
\midrule
Seq-1 & [758, 184, 20, 88, 184, 302, 88, 184] & 20 & 1.52 & 1.52 & 1.59 & 0.30 & 0.33 & 0.36 \\
Seq-2 & [1, 632, 1, 632, 1, 632, 1, 50] & 632 & 3.24 & 4.50 & 4.50 & 0.31 & 0.34 & 0.34 \\
Seq-3 & [3, 8, 3, 21, 3, 242, 3, 242] & 3  & 1.51 & 1.30 & 1.51 & 0.35 & 0.30 & 0.35 \\
Seq-4 & [57, 338, 519, 57, 338, 221, 57, 2] & 142   & 0.12 & 0.12 & 0.14 & 0.32 & 0.33 & 0.34 \\

\bottomrule
\end{tabular}
\label{tab:hop_pmi_alignment}
\end{table*}
\subsection{Case Study}
To complement the model-level interpretation, Section~S1 of the supplementary material first reports target-conditioned predecessor statistics, showing that 2-Hop and 3-Hop positions contain predictive signals beyond immediate 1-Hop adjacency. Building on this data-level evidence, we investigate whether the learned hop-weights align with the intrinsic structural dependencies of app transitions. Specifically, we conduct two groups of detailed analyses: (1) an alignment study between  hop-weights and structural correlations, (2) structured perturbation experiments to assess sensitivity and dependency of critical multi-hop structures.

\paragraph{Alignment of Hop-Weights and Structural Correlation}
To verify whether the learned hop-level attention weights ($W_{hop_{\gamma}}$) capture intrinsic dependencies in app transitions, we introduce a statistical benchmark based on Pointwise Mutual Information (PMI). Let $\tilde{V}^S_{\gamma} \subseteq V^S$ denote the set of non-isolated nodes in $G^S_{\gamma}$. Then,
\begin{equation}
sim_{\gamma} = \frac{1}{|\tilde{V}^S_{\gamma}|} \sum_{u \in \tilde{V}^S_{\gamma}} \log \frac{P(u,y)}{P(u)P(y)},
\end{equation}
where $P(u,y)$ denotes the joint probability of app $u$ and target app $y$ co-occurring in a session, while $P(u)$ and $P(y)$ denote their marginal probabilities. All probabilities are estimated from training-session statistics.

We analyze 50 samples where MISApp correctly predicts the target app but the variant using only the 1-Hop session graph predicts it incorrectly, and select those with the largest increases in target prediction probability after incorporating 2-Hop and 3-Hop structures. Alignment between learned hop attention and external transition statistics is measured using Kendall's Tau and the Top-1 Order Consistency Rate. The average Kendall's Tau is 0.71, with a Top-1 consistency rate of 0.78. The distribution in Table~\ref{tab:kendall_distribution} further indicates strong monotonic agreement in identifying the most influential hop.

\begin{figure}[htbp]
    \centering
    \includegraphics[width=1.0\columnwidth]{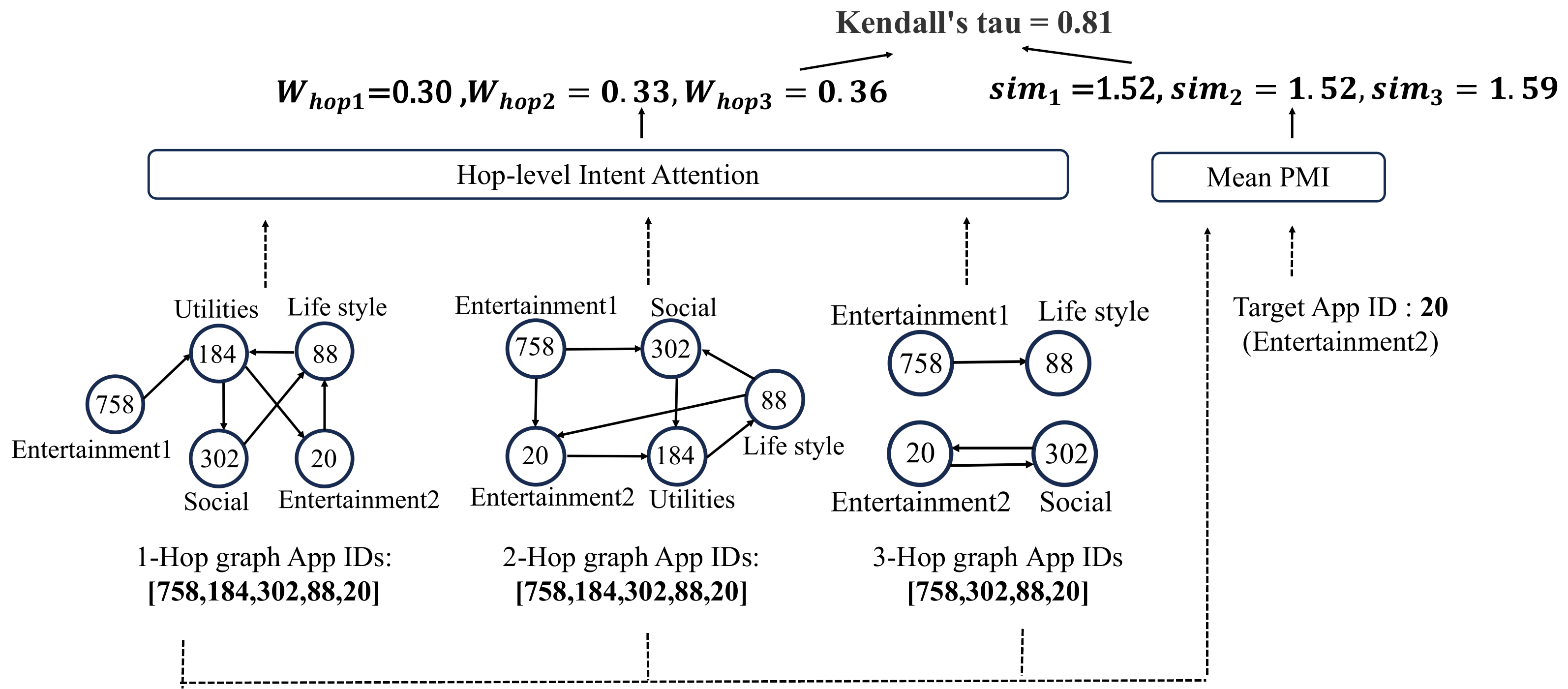}
    \caption{Hop-level weights and structural correlation mechanism for Seq-1. Each app is annotated with both the app ID and its corresponding category label.} 
    \label{fig: correlation mechanism}
\end{figure}

Representative examples are shown in Table~\ref{tab:hop_pmi_alignment}. For Seq-1, Fig.~\ref{fig: correlation mechanism} shows that the 3-Hop structure contains apps with stronger semantic relevance and behavioral alignment to the target category (i.e., Entertainment) than the 1-Hop and 2-Hop graphs. Multi-hop propagation suppresses incidental intermediate interactions while preserving intent-consistent transition paths, yielding a more discriminative structural context and explaining the higher attention assigned to the 3-Hop graph.

\paragraph{Structured Perturbation Experiment}
To further evaluate MISApp's sensitivity to critical structural information in multi-hop session graphs, we design a structured perturbation experiment based on the input sequences presented in Table~\ref{tab:hop_pmi_alignment}. This experiment investigates whether the model's prediction of the target app relies on high-relevance apps captured by the identified multi-hop structures. Specifically, the experiment consists of three steps: influential node identification, perturbation sequence construction, and model inference and comparison.

\begin{enumerate}
\item 
We first select the $\gamma$-Hop session graph assigned the largest hop weight $W_{hop_{\gamma}}$. Within this graph, the relevance between an app $v$ and the target app $y$ is measured by Pointwise Mutual Information (PMI):
\begin{equation}
\mathrm{PMI}(v,y)
=
\log
\frac{p(v,y)}
{p(v)p(y)},
\end{equation}
 A larger PMI indicates a stronger statistical association between the two apps. We then identify the app with the highest PMI with respect to $y$ in the selected $\gamma$-Hop graph as the influential node.

\item
A perturbed sequence is constructed by replacing the influential app $v$ with an alternative app $v^\ast$. To keep the perturbation contextually plausible and avoid introducing an unrelated replacement, $v^\ast$ is selected from the training sequence with the highest Jaccard similarity to the original input sequence. This strategy preserves the overall sequential context while altering the local app-level evidence associated with $v$.

\item
The perturbed sequence is fed into the model to recalculate the prediction probability $P(y)$. We compare this with the original probability to assess the model’s ability to capture high-order dependencies.
\end{enumerate}

\begin{figure}[!tbp]
    \centering
    \includegraphics[width=0.8\columnwidth]{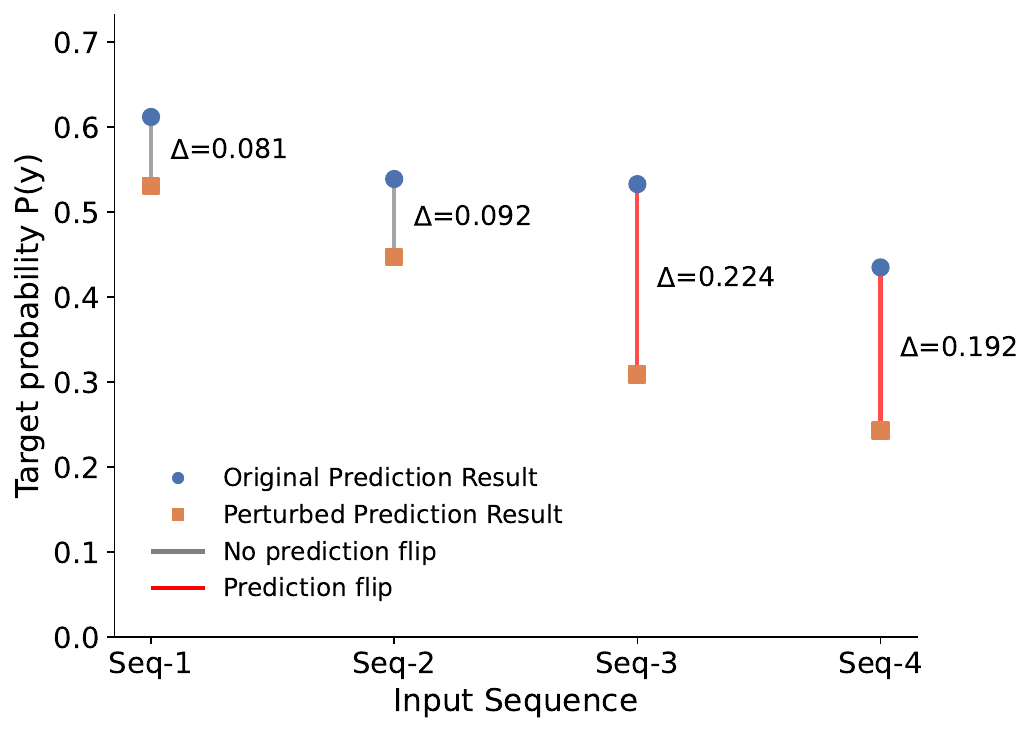}
    \caption{Comparison of Top-1 prediction results and target probabilities under original and perturbed structural settings. $\Delta$ denotes the drop in target prediction probability.}
    \label{fig:Perturbation}
\end{figure}
As observed in Fig.~\ref{fig:Perturbation}, under the Top-1 prediction setting, removing high-relevance apps from the maximum weight $\gamma$-Hop session graph leads to a substantial drop in the target probability. Among the illustrated samples, prediction flips occur for the samples with target probability drops greater than 0.1. This demonstrates that the removed apps carry critical contextual signals for correct inference, and further confirms that our multi-hop reasoning mechanism effectively captures procedural high-order dependencies inherent in user behavior sequences.


\section{CONCLUSION}
This paper studies next app prediction from the perspective of session-level structural dependency modeling and intent evolution. We show that explicitly modeling multi-hop transition structure provides a more informative view of app usage behavior than conventional sequential or local-transition formulations, especially when user-specific historical information is limited. Built on this idea, MISApp offers a profile-free solution that combines multi-hop session modeling with session-level intent inference. Experimental results on two real-world datasets demonstrate that this design yields strong performance under both standard and cold-start settings, while remaining practically efficient. Moreover, the learned hop-level attention weights also show close agreement with structural relevance, suggesting that the model distinguishes the predictive contributions of different hop ranges. Future work will focus on improving efficiency and scalability and on developing lighter variants for deployment in resource-constrained mobile environments.

\clearpage
\section*{Supplementary Material}
\setcounter{section}{0}
\setcounter{subsection}{0}
\setcounter{figure}{0}
\setcounter{table}{0}
\setcounter{equation}{0}
\renewcommand{\thesection}{S\arabic{section}}
\renewcommand{\thesubsection}{\thesection.\arabic{subsection}}
\renewcommand{\thefigure}{S\arabic{figure}}
\renewcommand{\thetable}{S\arabic{table}}
\renewcommand{\theequation}{S\arabic{equation}}
\renewcommand{\theHsection}{supp.section.\arabic{section}}
\renewcommand{\theHsubsection}{supp.subsection.\arabic{section}.\arabic{subsection}}
\renewcommand{\theHfigure}{supp.figure.\arabic{figure}}
\renewcommand{\theHtable}{supp.table.\arabic{table}}
\renewcommand{\theHequation}{supp.equation.\arabic{equation}}
This document provides the additional structural and spatial analyses referenced in the main paper. Section~S1 analyzes multi-hop dependencies, Section~S2 compares alternative graph and hypergraph formulations, Section~S3 studies spatial context construction, and Section~S4 reports parameter sensitivity analyses.
\section{Multi-hop Dependency Analysis}
\label{sec:multi_hop_dependency_analysis}

To better justify the multi-hop session graph design in MISApp, we analyze multi-hop dependencies from two complementary perspectives. First, we conduct a target-conditioned predecessor analysis to examine whether higher-hop positions contain predictive signals beyond direct 1-Hop adjacency at the data level. Second, we perform a hop combination study to verify whether these multi-range structural signals can effectively improve next app prediction at the model level.

\paragraph{Target-conditioned predecessor analysis}
For each target app $y$, we collect all samples whose target app is $y$.
Given an input sequence $(a_1,a_2,\dots,a_T)$ and the target app
$a_{T+1}=y$, we estimate the empirical probability that app $x$ appears
at the $l$-th position before $y$:
\begin{equation}
p_l(x|y)=
\frac{
\mathrm{Count}(a_{T+1-l}=x, a_{T+1}=y)
}{
\sum_{x'\in\mathcal{A}}
\mathrm{Count}(a_{T+1-l}=x', a_{T+1}=y)
},
\end{equation}
where $\mathrm{Count}(\cdot)$ denotes the number of samples satisfying
the specified condition. Accordingly, $p_l(x|y)$ represents the empirical
conditional probability that app $x$ appears at the $l$-th position before
the target app $y$. Here, $l\in\{1,2,3\}$ corresponds to the last-1,
last-2, and last-3 positions before the target, respectively.

For each position, we define the Top-3 predecessor set as
\begin{equation}
\mathcal{T}_l^3(y)=\mathrm{Top}_3(p_l(\cdot|y)).
\end{equation}
The Top-3 mass is then defined as
\begin{equation}
M_l^3(y)=
\sum_{x\in\mathcal{T}_l^3(y)}
p_l(x|y),
\end{equation}
which measures whether the predecessor distribution at position $l$ is
dominated by a small number of apps. A larger $M_l^3(y)$ indicates a more
concentrated and stable dependency pattern at the corresponding position.

To quantify whether higher-hop positions provide information beyond
immediate 1-Hop adjacency, we further define the Top-3 unique mass:
\begin{equation}
U_l^3(y)=
\sum_{x\in\mathcal{T}_l^3(y)\setminus \mathcal{T}_1^3(y)}
p_l(x|y),
\quad l\in\{2,3\}.
\end{equation}
A larger $U_l^3(y)$ indicates that the dominant predecessors at the
$l$-Hop position are not covered by the most frequent 1-Hop predecessors,
suggesting higher-hop-specific predictive signals.

Table~\ref{tab:target_position_cases} presents representative target apps from different frequency ranges, together with the dominant predecessors and the corresponding Top-3 mass and unique mass at different positions. The results show that the dominant apps at the last-2 and last-3 positions often differ from those at the last-1 position while still exhibiting strong dependency patterns and substantial unique mass. This indicates that higher-hop positions provide complementary predictive signals beyond direct 1-Hop adjacency, capturing delayed revisit and functional alternation patterns that cannot be fully recovered from immediate transitions alone.

\paragraph{Hop combination study}
Based on the above data-level observations, we further evaluate whether
different hop ranges provide complementary predictive benefits in MISApp.
Specifically, we compare single-hop variants, two-hop combinations, and
the full 1-Hop + 2-Hop + 3-Hop setting on both datasets under the standard
split. The results are shown in Table~\ref{tab:hop_comparison}.

First, different single-hop graphs exhibit distinct predictive strengths.
On the Tsinghua App Usage dataset, the 2-Hop graph performs better than
the 1-Hop graph on several metrics, while on LSapp, the 2-Hop and 3-Hop
graphs also show clear advantages over the 1-Hop graph. This observation
is consistent with the preceding data-level analysis, which shows that
higher-hop positions contain predictive information beyond immediate
1-Hop adjacency. These results further suggest that non-consecutive app
transitions capture useful behavioral dependencies that cannot be fully
represented by direct transitions alone.

Second, combining multiple hop ranges generally yields stronger
performance than relying on a single hop, indicating that different
structural ranges provide complementary rather than redundant information.
Among these configurations, the full 1-Hop + 2-Hop + 3-Hop setting
achieves the best overall performance on both datasets, demonstrating the
benefit of jointly modeling direct, intermediate, and higher-order
transition dependencies.

\begin{table*}[t]
\centering
\caption{Representative examples of target-conditioned predecessor distributions on  LSapp. 
Last-$i$ denotes the app appearing at the $i$-th position before the target app, and the value in parentheses denotes the empirical probability $p_i(x|y)$.}
\label{tab:target_position_cases}
\renewcommand{\arraystretch}{1.3} 
\setlength{\tabcolsep}{6pt} 
\begin{tabular}{l|ccc|ccc|cc}
\toprule
Target App
& Last-1 Top App (Prob)
& Last-2 Top App (Prob)
& Last-3 Top App (Prob)
& $M_1^3$
& $M_2^3$
& $M_3^3$
& $U_2^3$
& $U_3^3$ \\
\midrule
Contacts
& Android In Call UI (0.424)
& Contacts (0.755)
& Android In Call UI (0.393)
& 0.623 & 0.800 & 0.574 & 0.800 & 0.074 \\

Movie Play Box
& Reward Stash (0.776)
& Movie Play Box (0.881)
& Reward Stash (0.735)
& 0.904 & 0.955 & 0.869 & 0.914 & 0.000 \\

Army Men Strike
& Yahoo Mail (0.219)
& Army Men Strike (0.642)
& Army Men Strike (0.175)
& 0.511 & 0.818 & 0.467 & 0.642 & 0.175 \\
\bottomrule
\end{tabular}
\end{table*}

\begin{table*}[htbp]
    \centering
   
    \caption{Performance comparison of different hop combinations under the standard split. The best results are highlighted in bold, and the second-best results are underlined.} 
    \label{tab:hop_comparison}

        \resizebox{\textwidth}{!}{%
        \begin{tabular}{lcccccccccc}
            \toprule
            \multirow{2}{*}{Hop Combination} 
            & \multicolumn{5}{c}{Tsinghua App Usage} 
            & \multicolumn{5}{c}{LSapp} \\
            \cmidrule(lr){2-6} \cmidrule(lr){7-11}
            & ACC@1 & ACC@3 & ACC@5 & MRR@3 & MRR@5 
            & ACC@1 & ACC@3 & ACC@5 & MRR@3 & MRR@5 \\
            \midrule
        
            1-Hop only 
            & 51.56$_{\pm0.062}$ & 71.92$_{\pm0.045}$ & 80.91$_{\pm0.019}$ & 62.02$_{\pm0.048}$ & 63.86$_{\pm0.044}$ 
            & 51.82$_{\pm0.20}$ & 90.90$_{\pm0.019}$ & 94.13$_{\pm0.043}$ & 70.47$_{\pm0.100}$ & 71.24$_{\pm0.095}$ \\
            
            2-Hop only 
            & 53.84$_{\pm0.054}$ & 75.36$_{\pm0.061}$ & 80.61$_{\pm0.084}$ & 63.43$_{\pm0.052}$ & 65.10$_{\pm0.055}$ 
            & \underline{83.90}$_{\pm0.024}$ & 91.90$_{\pm0.095}$ & 94.17$_{\pm0.108}$ & 87.54$_{\pm0.052}$ & 88.06$_{\pm0.051}$ \\
            
            3-Hop only 
            & 53.51$_{\pm0.208}$ & 75.05$_{\pm0.254}$ & 81.47$_{\pm0.177}$ & 63.12$_{\pm0.226}$ & 64.82$_{\pm0.209}$ 
            & 80.03$_{\pm0.150}$ & 91.42$_{\pm0.084}$ & 94.24$_{\pm0.048}$ & 82.14$_{\pm0.126}$ & 85.79$_{\pm0.116}$ \\
            
            1-Hop + 2-Hop 
            & \underline{54.34}$_{\pm0.115}$ & 75.95$_{\pm0.166}$ & \underline{82.16}$_{\pm0.155}$ & \underline{63.99}$_{\pm0.139}$ & \underline{65.65}$_{\pm0.136}$ 
            & 83.75$_{\pm0.025}$ & 92.15$_{\pm0.147}$ & 94.34$_{\pm0.159}$ & 84.57$_{\pm0.069}$ & 88.07$_{\pm0.071}$ \\
            
            1-Hop + 3-Hop 
            & 53.69$_{\pm0.119}$ & 75.30$_{\pm0.175}$ & 81.75$_{\pm0.158}$ & 63.31$_{\pm0.145}$ & 65.03$_{\pm0.140}$ 
            & 80.66$_{\pm0.329}$ & 91.34$_{\pm0.081}$ & 94.23$_{\pm0.017}$ & 85.43$_{\pm0.218}$ & 86.09$_{\pm0.202}$ \\
            
            2-Hop + 3-Hop 
            & 54.26$_{\pm0.101}$ & \underline{75.97}$_{\pm0.173}$ & 82.13$_{\pm0.126}$ & 63.94$_{\pm0.132}$ & 65.59$_{\pm0.122}$ 
            & 83.73$_{\pm0.025}$ & \underline{92.39}$_{\pm0.206}$ & \underline{94.66}$_{\pm0.203}$ & \underline{87.67}$_{\pm0.113}$ & \underline{88.19}$_{\pm0.113}$ \\
            
            MISApp 
            & \textbf{54.52$_{\pm0.175}$} & \textbf{76.33$_{\pm0.271}$} & \textbf{82.40$_{\pm0.240}$} & \textbf{64.26$_{\pm0.214}$} & \textbf{65.89$_{\pm0.209}$} 
            & \textbf{84.00$_{\pm0.038}$} & \textbf{92.44$_{\pm0.167}$} & \textbf{94.87$_{\pm0.142}$} & \textbf{87.93$_{\pm0.095}$} & \textbf{88.63$_{\pm0.090}$} \\
            
            \bottomrule
        \end{tabular}
    }
\end{table*}

\section{Comparison with Conventional Graph and Hypergraph Modeling}

To further clarify the difference between MISApp and conventional graph- or hypergraph-based structural modeling, we conduct a controlled comparison by replacing only the structural modeling module while keeping the remaining architecture unchanged. For fair comparison, the graph neural network depth of Multi-hop GCN, Session Hypergraph, Merged Multi-hop Graph, and Multi-hop w/o Hop Attention is set to $L_g=2$, consistent with MISApp.

Specifically, 1-Hop 1-layer GCN applies a 1-layer GCN to the 1-Hop session graph, while 1-Hop 3-layer GCN applies a deeper 3-layer GCN to the same graph to examine whether increasing propagation depth can implicitly capture higher-order dependencies. Multi-hop GCN separately constructs 1-Hop, 2-Hop, and 3-Hop graphs and applies a 2-layer GCN to each hop-specific graph. Session Hypergraph treats the 1-Hop, 2-Hop, and 3-Hop paths as different types of hyperedges and employs a 2-layer hypergraph neural network to learn session representations. Merged Multi-hop Graph merges the edges from the 1-Hop, 2-Hop, and 3-Hop graphs into a single graph and applies LightGCN to the merged graph. Multi-hop w/o Hop Attention retains the multi-hop LightGCN structure of MISApp with $L_g=2$, but removes the attention-based fusion over hop-specific graph embeddings. The results are reported in Table~\ref{tab:structural_modeling}.

The results show that conventional GCN-based variants are not sufficient to fully capture effective session structures. Although stacking multiple GCN layers can implicitly expand the receptive field, the 1-Hop 3-layer GCN performs worse than the 1-Hop 1-layer GCN on both datasets. This indicates that simply increasing propagation depth may introduce noisy or over-smoothed representations, rather than producing more discriminative higher-order dependencies. The Multi-hop GCN improves performance on LSapp, suggesting that higher-order transition information is useful; however, it still lags behind MISApp because it lacks an explicit mechanism to adaptively distinguish and integrate the contributions of different hop-specific structural representations.

The Session Hypergraph variant also performs poorly, especially on the Tsinghua App Usage dataset. This is because hypergraph modeling mainly captures high-order co-occurrence among apps within a session, but weakens the sequential distance between app transitions. In contrast, MISApp preserves the distinction among 1-Hop, 2-Hop, and 3-Hop dependencies, which is important for modeling direct transitions, intermediate associations, and moderate-range behavioral routines.

Moreover, the Merged Multi-hop Graph and Multi-hop w/o Hop Attention variants achieve stronger results than conventional GCN and hypergraph variants, confirming the usefulness of multi-hop structural information. The Merged Multi-hop Graph uses the same 1-Hop, 2-Hop, and 3-Hop edges but collapses them into a single graph, thereby losing hop-specific structural semantics. Multi-hop w/o Hop Attention keeps separated hop graphs but fuses them uniformly, ignoring that different sessions may rely on different structural ranges.

\begin{table}[t]
\centering
\caption{Comparison with conventional graph and hypergraph structural modeling under the standard split. The best results are highlighted in bold, and the second-best results are underlined.}
\label{tab:structural_modeling}
\footnotesize 
\setlength{\tabcolsep}{2.5pt} 
\renewcommand{\arraystretch}{0.85} 
\scalebox{0.9}{ 
\begin{tabular}{l|cc|cc}
\toprule
\multirow{2}{*}{Structural Module} 
& \multicolumn{2}{c|}{Tsinghua App Usage} 
& \multicolumn{2}{c}{LSapp} \\
\cmidrule(lr){2-3} \cmidrule(lr){4-5}
& ACC@1 & MRR@3 & ACC@1 & MRR@3 \\
\midrule
1-Hop 1-layer GCN & 50.68$_{\pm0.043}$ & 60.05$_{\pm0.059}$ & 52.46$_{\pm0.073}$ & 70.77$_{\pm0.047}$ \\
1-Hop 3-layer GCN & 46.14$_{\pm0.062}$ & 56.28$_{\pm0.050}$ & 51.96$_{\pm0.146}$ & 70.40$_{\pm0.075}$ \\
Multi-hop GCN & 47.75$_{\pm0.097}$ & 57.52$_{\pm0.075}$ & 80.45$_{\pm0.055}$ & 84.57$_{\pm0.034}$ \\
Session Hypergraph & 44.97$_{\pm0.062}$ & 57.10$_{\pm0.071}$ & 55.90$_{\pm0.066}$ & 72.77$_{\pm0.041}$ \\
Merged Multi-hop Graph & 52.44$_{\pm0.086}$ & 62.18$_{\pm0.085}$ & \underline{80.90}$_{\pm0.021}$ & \underline{84.97}$_{\pm0.047}$ \\
Multi-hop w/o Hop Att & \underline{52.65}$_{\pm0.060}$ & \underline{62.37}$_{\pm0.078}$ & 80.59$_{\pm0.022}$ & 84.56$_{\pm0.014}$ \\
\midrule
MISApp & \textbf{54.52$_{\pm0.175}$} & \textbf{64.26$_{\pm0.214}$} & \textbf{84.00$_{\pm0.038}$} & \textbf{87.93$_{\pm0.095}$} \\
\bottomrule
\end{tabular}
}
\end{table}

\section{Spatial Context Construction Study}
\label{sec:spatial_context_study}
Unlike conventional spatial modeling approaches that mainly rely on
geographical proximity or discrete location identifiers, our spatial
context construction aims to capture the functional similarity among
locations. In mobile app usage scenarios, two geographically distant
locations may share similar user behaviors if they have similar
functional characteristics, while nearby locations may correspond to
different usage patterns. Therefore, directly modeling spatial distance
or base station IDs may fail to capture the functional relationships
embedded in POI attributes. 

To address this issue, we construct a top-$k$ POI similarity graph, where
base stations are connected according to their POI feature similarity.
The connected components of this graph are then regarded as spatial
categories, enabling the model to capture groups of locations with
similar functional contexts. The spatial neighbor number $k_{loc}$ 
controls the granularity of the similarity graph: a small value may
produce fragmented spatial categories with insufficient connections,
whereas a large value may introduce weakly related locations and reduce
category discriminability.

\begin{figure*}[t]
\centering
\includegraphics[width=0.8\textwidth]{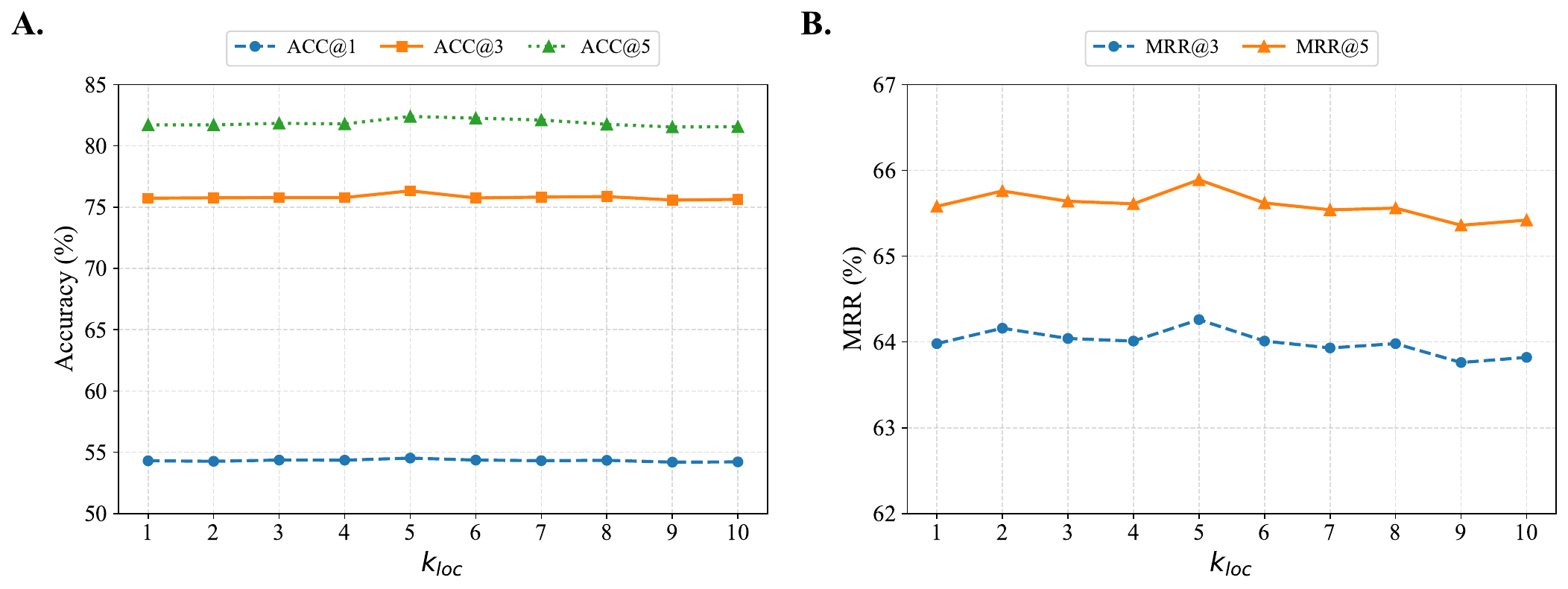}
\caption{Sensitivity analysis of the spatial neighbor number $k_{loc}$ on the Tsinghua App Usage  under the standard split. A. Accuracy-based evaluation results, including ACC@1, ACC@3, and ACC@5. 
B. Ranking-based evaluation results, including MRR@3 and MRR@5.}
\label{fig:kloc_sensitivity}
\end{figure*}

We first examine the effect of the spatial neighbor number $k_{loc}$ in
the proposed top-$k$ POI similarity graph. As shown in
Fig.~\ref{fig:kloc_sensitivity}, the prediction performance varies with
different values of $k_{loc}$, and the best overall performance is
achieved when $k_{loc}=5$. When $k_{loc}$ becomes larger, the performance
slightly decreases, suggesting that excessive connections may introduce
weakly related base stations into the spatial graph and weaken the
functional distinction among spatial categories. Therefore, we set
$k_{loc}=5$ in the following experiments.

After determining $k_{\mathrm{loc}}$, we further compare the proposed
spatial context construction strategy with several representative
alternatives on the Tsinghua App Usage dataset. In addition to directly
using learnable base station ID embeddings, we evaluate four POI
clustering-based methods: K-means~\cite{kmeans},
K-means++~\cite{k-means++}, K-Harmonic Means~\cite{k-harmonic},
and K-modes~\cite{kmodes}. Let $K_c$ denote the number of clusters,
$z_r\in\{1,\dots,K_c\}$ denote the cluster assignment of base station
$r$, and $\boldsymbol{\mu}_j$ denote the prototype of the $j$-th cluster.

K-means partitions base stations by minimizing the within-cluster
squared Euclidean distance:
\begin{equation}
\mathcal{L}_{\mathrm{KM}}
=
\sum_{r\in\mathcal{R}}
\left\|
\tilde{\mathbf{x}}_r-\boldsymbol{\mu}_{z_r}
\right\|_2^2 ,
\end{equation}
where $z_r$ denotes the cluster assignment of base station $r$, and
$\boldsymbol{\mu}_{z_r}$ is the centroid of its assigned cluster.

K-means++ uses the same clustering objective as K-means but adopts a
distance-aware strategy to initialize the cluster centers. Given the set
$\mathcal{M}$ of centers that have already been selected, the probability
of choosing base station $r$ as the next center is
\begin{equation}
P(r)
=
\frac{D(r)^2}
{\sum_{q\in\mathcal{R}}D(q)^2},
\quad
D(r)=
\min_{\boldsymbol{\mu}\in\mathcal{M}}
\left\|
\tilde{\mathbf{x}}_r-\boldsymbol{\mu}
\right\|_2 ,
\end{equation}
where $D(r)$ is the distance from base station $r$ to its nearest
previously selected center. Thus, base stations farther from the existing
centers are more likely to be selected as new initial centers.

K-Harmonic Means considers the distances from each base station to all
cluster centers rather than only its nearest center, using the following
harmonic-distance-based objective:
\begin{equation}
\mathcal{L}_{\mathrm{KHM}}
=
\sum_{r\in\mathcal{R}}
\frac{K_c}
{\sum_{j=1}^{K_c}
\left(
\left\|
\tilde{\mathbf{x}}_r-\boldsymbol{\mu}_j
\right\|_2^2+\epsilon
\right)^{-1}},
\end{equation}
where $\epsilon$ is a small constant for numerical stability.

K-modes replaces numerical centroids with mode-based cluster
representatives. For cluster $j$, let
$\boldsymbol{\theta}_j=(\theta_{j,1},\ldots,\theta_{j,M})$ denote its mode
vector, where $\theta_{j,m}$ is the most frequent value of the $m$-th
feature among the base stations assigned to cluster $j$. The clustering
objective minimizes the number of feature-wise mismatches between each
base station and its assigned mode vector:
\begin{equation}
\mathcal{L}_{\mathrm{KMD}}
=
\sum_{r\in\mathcal{R}}
\sum_{m=1}^{M}
\mathbb{I}
\left(
\tilde{x}_{r,m}\neq\theta_{z_r,m}
\right),
\end{equation}
where $\mathbb{I}(\cdot)$ equals 1 when the two feature values are
different and 0 otherwise.

For the clustering-based baselines, the number of clusters is selected
by the elbow method according to the corresponding clustering objective.
Specifically, the optimal number of clusters is 3 for K-means, and 4 for
K-means++, K-Harmonic Means, and K-modes. In contrast, our method does
not rely on a predefined number of prototypes. It constructs a top-$k_{\mathrm{loc}}$
POI similarity graph and obtains 8 spatial categories as connected
components. All variants share the same backbone architecture, and only
the spatial context construction module is changed.

The results are reported in Table~\ref{tab:spatial_construction}. Direct base station ID embeddings learn each base station independently, making it difficult to obtain reliable representations for low-frequency base stations with limited observations. Clustering-based methods provide compact spatial abstractions by grouping base stations according to POI features. However, these methods mainly rely on distance-based assignments, which may be less suitable for heterogeneous base-station environments.

To further examine whether the constructed spatial categories preserve
functional consistency, we compare the intra-category and inter-category
POI similarities of different spatial construction methods. For each spatial construction method, 
let $\mathcal{C}=\{C_1,C_2,\dots,C_F\}$ denote the generated spatial
categories, where $F$ is the number of categories. We compute the
intra-category similarity as
\begin{equation}
\mathrm{IntraSim}
=
\frac{1}{|\mathcal{F}|}
\sum_{f\in\mathcal{F}}
\frac{2}{|C_f|(|C_f|-1)}
\sum_{\substack{r,q\in C_f}}
\mathrm{sim}_{\mathrm{loc}}(r,q),
\end{equation}
where $\mathcal{F}=\{f\mid |C_f|>1\}$ denotes the set of
categories containing at least two base stations.

The inter-category similarity is computed by averaging the
similarity over all pairs of different categories:
\begin{equation}
S_{fg}
=
\frac{1}{|C_f||C_g|}
\sum_{r\in C_f}
\sum_{q\in C_g}
\mathrm{sim}_{\mathrm{loc}}(r,q),
\end{equation}

\begin{equation}
\mathrm{InterSim}
=
\frac{2}{F(F-1)}
\sum_{1\leq f<g\leq F}
S_{fg}.
\end{equation}
The similarity gap is defined as
\begin{equation}
\mathrm{Gap}_{\mathrm{sim}}=\mathrm{IntraSim}-\mathrm{InterSim}.
\end{equation}
A larger positive $\mathrm{Gap}_{\mathrm{sim}}$ indicates that base
stations assigned to the same spatial category have more similar POI
compositions than those assigned to different categories.

We additionally evaluate category separation from the complementary
perspective of POI-feature distance. We first obtain the unit-normalized POI representation by
\begin{equation}
\hat{\mathbf{x}}_r
=
\frac{\tilde{\mathbf{x}}_r}{\|\tilde{\mathbf{x}}_r\|_2},
\end{equation}
and define the normalized POI Euclidean distance between two base stations as
\begin{equation}
\mathrm{dist}_{\mathrm{loc}}(r,q)
=
\left\|\hat{\mathbf{x}}_r-\hat{\mathbf{x}}_q\right\|_2.
\end{equation}
Following the same class-wise macro-averaging protocol, the intra-category
distance is
\begin{equation}
\mathrm{IntraDist}
=
\frac{1}{|\mathcal{F}|}
\sum_{f\in\mathcal{F}}
\frac{2}{|C_f|(|C_f|-1)}
\sum_{\substack{r,q\in C_f}}
\mathrm{dist}_{\mathrm{loc}}(r,q).
\end{equation}
For each pair of different categories, we compute
\begin{equation}
D_{fg}
=
\frac{1}{|C_f||C_g|}
\sum_{r\in C_f}
\sum_{q\in C_g}
\mathrm{dist}_{\mathrm{loc}}(r,q),
\end{equation}
and then average over all category pairs:
\begin{equation}
\mathrm{InterDist}
=
\frac{2}{F(F-1)}
\sum_{1\leq f<g\leq F}D_{fg}.
\end{equation}
Unlike similarity, desirable spatial categories should have a small
$\mathrm{IntraDist}$ and a large $\mathrm{InterDist}$. We therefore define
the distance gap in the opposite direction:
\begin{equation}
\mathrm{Gap}_{\mathrm{dist}}
=
\mathrm{InterDist}-\mathrm{IntraDist}.
\end{equation}
A larger positive $\mathrm{Gap}_{\mathrm{dist}}$ indicates stronger
separation between spatial categories.

\begin{figure*}[t]
\centering
\includegraphics[width=0.98\textwidth]{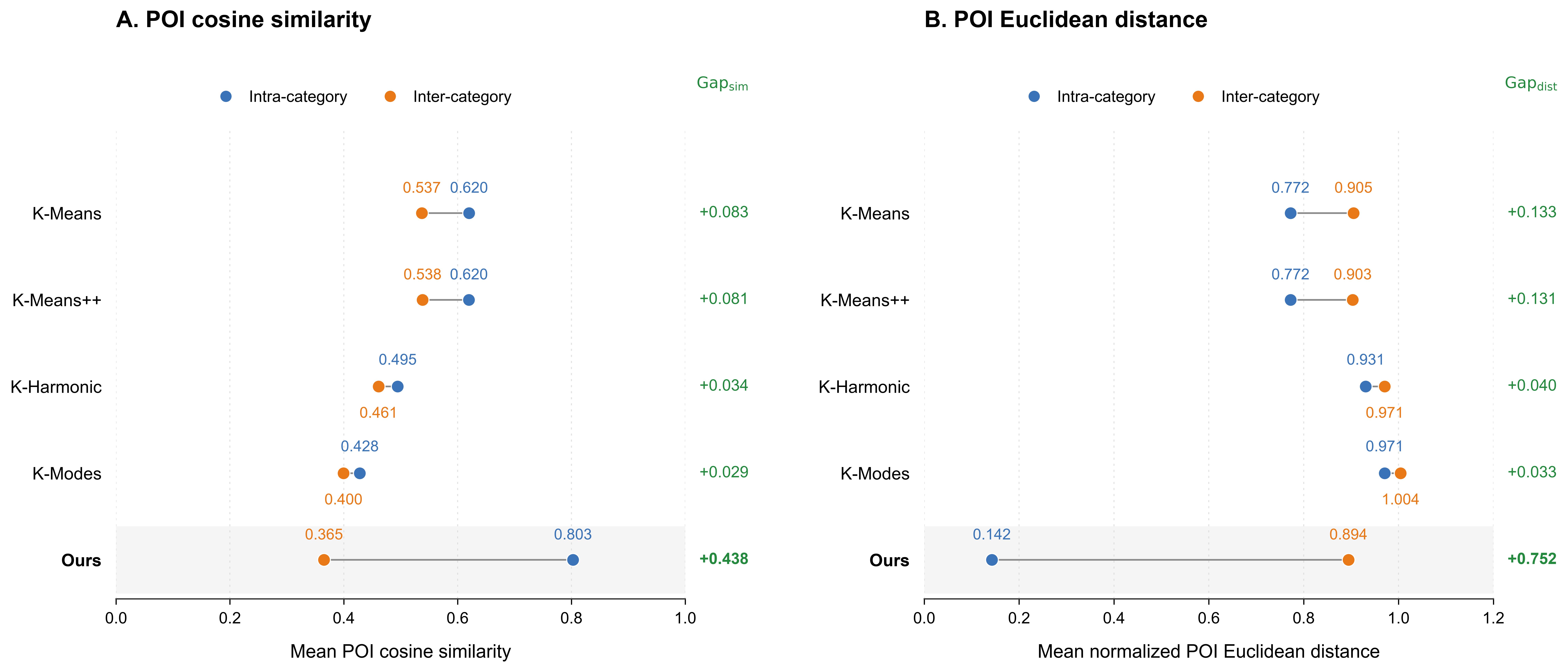}
\caption{Class-wise comparison of POI consistency under different spatial
context construction methods. A. Mean intra-category and inter-category
cosine similarities, where
$\mathrm{Gap}_{\mathrm{sim}}=\mathrm{IntraSim}-\mathrm{InterSim}$.
B. Mean intra-category and inter-category normalized Euclidean distances,
where
$\mathrm{Gap}_{\mathrm{dist}}=\mathrm{InterDist}-\mathrm{IntraDist}$.
For both panels, a larger positive gap indicates more compact and better
separated spatial categories.}
\label{fig:spatial_similarity_gap}
\end{figure*}

As shown in Fig.~\ref{fig:spatial_similarity_gap}, the proposed method achieves stronger category separation than the competing strategies under both similarity- and distance-based criteria. The consistent improvements indicate that the similarity-graph connected-component strategy produces compact and well-separated spatial categories. These categories provide coherent spatial context for next app prediction.

\begin{table}[t]
\centering
\caption{Comparison of spatial context construction strategies on Tsinghua App Usage. The best results are highlighted in bold, and the second-best results are underlined.}
\label{tab:spatial_construction}
\footnotesize
\scriptsize
\setlength{\tabcolsep}{2pt}
\renewcommand{\arraystretch}{0.9}
\begin{tabular}{l|ccccc}
\toprule
Strategy & ACC@1 & ACC@3 & ACC@5 & MRR@3 & MRR@5 \\
\midrule
BS ID Emb. & 53.07$_{\pm0.075}$ & 74.53$_{\pm0.147}$ & 81.45$_{\pm0.117}$ & 62.63$_{\pm0.110}$ & 63.22$_{\pm0.102}$ \\
K-means & \underline{53.71}$_{\pm0.037}$ & \underline{75.20}$_{\pm0.283}$ & \underline{82.25}$_{\pm0.032}$ & \underline{63.66}$_{\pm0.048}$ & \underline{64.78}$_{\pm0.038}$ \\
K-means++ & 53.68$_{\pm0.047}$ & 74.93$_{\pm0.064}$ & 81.88$_{\pm0.081}$ & 63.06$_{\pm0.041}$ & 64.55$_{\pm0.043}$ \\
K-Harmonic & 53.45$_{\pm0.032}$ & 74.73$_{\pm0.104}$ & 81.73$_{\pm0.051}$ & 62.96$_{\pm0.062}$ & 64.56$_{\pm0.046}$ \\
K-modes & 53.38$_{\pm0.028}$ & 74.92$_{\pm0.061}$ & 81.66$_{\pm0.042}$ & 62.90$_{\pm0.023}$ & 63.70$_{\pm0.017}$ \\
\midrule
Ours & \textbf{54.52$_{\pm0.175}$} & \textbf{76.33$_{\pm0.271}$} & \textbf{82.40$_{\pm0.240}$} & \textbf{64.26$_{\pm0.214}$} & \textbf{65.89$_{\pm0.209}$}  \\
\bottomrule
\end{tabular}
\end{table}
\section{Parameter Sensitivity Analysis}
To further assess the stability and generalization capability of MISApp, we conduct robustness experiments on the Tsinghua App Usage dataset under the standard split by varying several key architectural choices and hyperparameter settings. In particular, we examine how the immediate intent window size, embedding dimension, maximum session window size, and number of Transformer layers affect prediction performance.

\begin{figure*}[t]
    \centering
    \includegraphics[width=0.8\textwidth]{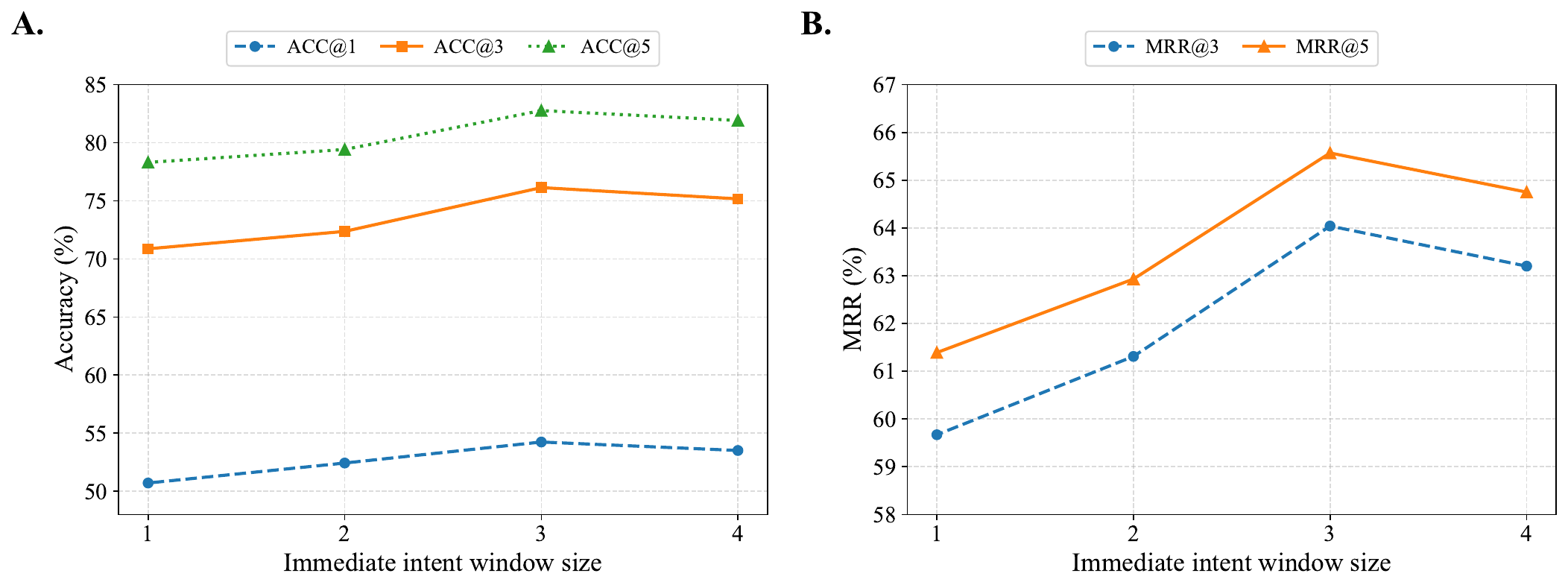} \\
    \caption{Effect of immediate intent window size on next app prediction performance on the Tsinghua App Usage dataset under the standard split.
A. Accuracy-based evaluation results, including ACC@1, ACC@3, and ACC@5. 
B. Ranking-based evaluation results, including MRR@3 and MRR@5.}
\label{fig:intent_window_size}
    \label{fig:last_app}

    \vspace{1em} 

    \includegraphics[width=0.8\textwidth]{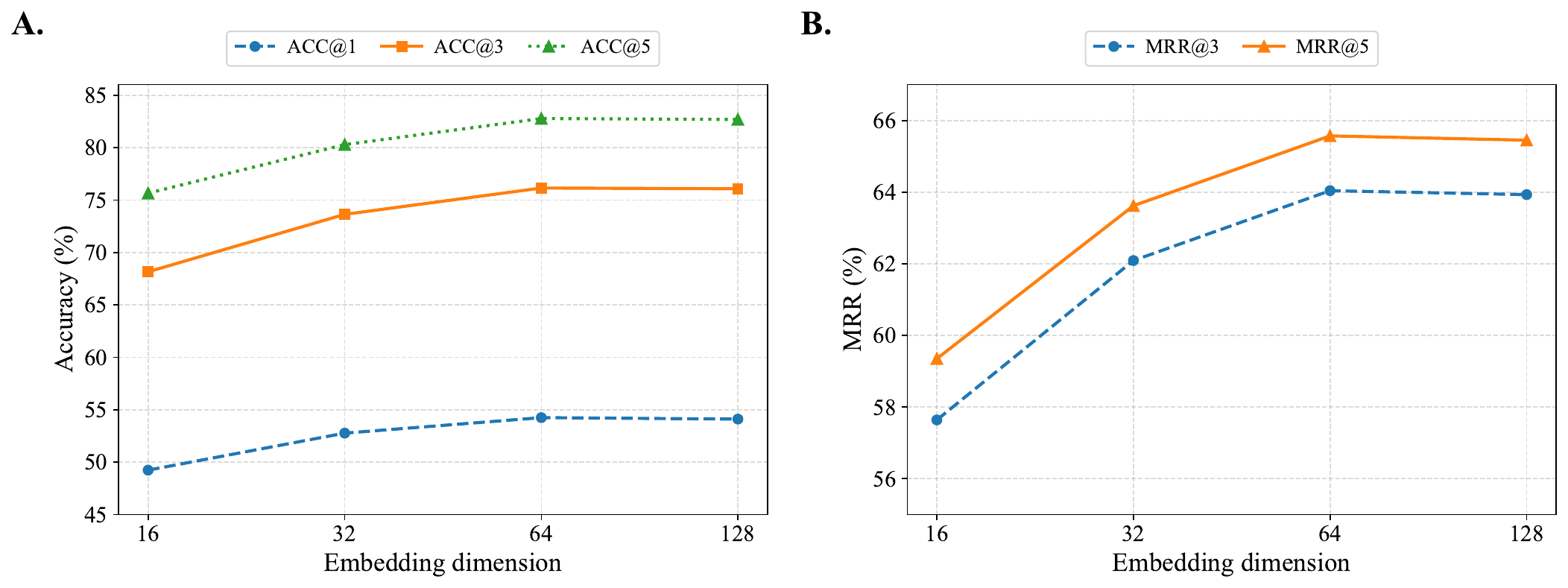} \\
    \caption{Effect of embedding dimension on next app prediction performance on the Tsinghua App Usage dataset under the standard split. 
A. Accuracy-based evaluation results, including ACC@1, ACC@3, and ACC@5. 
B. Ranking-based evaluation results, including MRR@3 and MRR@5.}
\label{fig:embedding_dimension}
    \label{fig:dim}
\end{figure*}

\paragraph{Effectiveness of Immediate Intent Window Size}
The immediate intent window size controls how many of the most recent app interactions are used to characterize short-term intent. We experiment with $K \in \{1, 2, 3, 4\}$ and observe that small to moderate window sizes consistently improve performance, highlighting the strong predictive value of short-term behavior. As shown in Fig.~\ref{fig:last_app}, when $K$ becomes excessively large, performance plateaus or slightly declines due to the inclusion of less relevant interactions. 

\paragraph{Effectiveness of Embedding Dimension}
We further analyze the effect of the embedding dimension $d$ on model
performance and computational cost. Let $T$, $\Gamma$, $L_g$, $L$,
$|\mathcal{A}|$, and $F$ denote the session window size, maximum hop
number, number of LightGCN layers, number of Transformer layers,
application vocabulary size, and number of spatial categories,
respectively. The main computational cost of MISApp comes from
multi-hop graph propagation, graph-level attention, Transformer-based
intent modeling, cross-modal fusion, and full-ranking prediction.

For the $\Gamma$ hop-specific session graphs, LightGCN propagation
requires $O(\Gamma L_g T d)$, while graph-level attention introduces
$O(\Gamma T d^2)$ due to the projection and aggregation of graph-node
representations. The Transformer encoder and decoder operate on the
fused session representation and immediate-intent representation. Therefore, the Transformer-based
intent modeling introduces $O(Ld^2)$ complexity. The cross-modal
fusion and immediate-intent projection introduce additional
$O(d^2)$ costs, which are of the same order as the other projection
operations. Finally, full-ranking prediction over all candidate apps
requires $O(d|\mathcal{A}|)$.

Accordingly, the overall per-session time complexity can be summarized as
\begin{equation}
\begin{aligned}
O(&\Gamma L_gTd+\Gamma Td^2 \\
  &+Ld^2+d|\mathcal{A}|).
\end{aligned}
\end{equation}

From the above expression, the computational cost grows linearly with
the maximum hop number $\Gamma$ and the session window size $T$, while
the full-ranking cost grows linearly with the application vocabulary
size $|\mathcal{A}|$. The embedding dimension $d$ has a quadratic
effect on the projection and feature-interaction operations. Since
$\Gamma$, $L_g$, and $L$ are small constants in our setting, the
dominant per-session time complexity can be further simplified as
\begin{equation}
O(Td^2+d|\mathcal{A}|).
\end{equation}

The online space complexity mainly consists of the embedding tables,
hop-specific graph representations, Transformer representations and
parameters, and full-ranking prediction scores, and can be summarized as
\begin{equation}
O((|\mathcal{A}|+F)d+\Gamma Td+Ld^2+|\mathcal{A}|).
\end{equation}
Here, the fixed-size temporal embedding table and other $O(d^2)$
projection parameters are absorbed into the corresponding lower-order
terms. Both the time and space complexities indicate that increasing
$d$ directly increases the costs of feature representation, projection,
and candidate scoring.

As shown in Fig.~\ref{fig:dim}, although larger embedding dimensions
generally improve representation capacity, the performance gain from
$d=64$ to $d=128$ is marginal. Hence, we set $d=64$ in MISApp to
achieve a better balance between prediction accuracy and computational
efficiency.

\paragraph{Effectiveness of Maximum Session Window Size} 
We vary the session window size $T \in \{5, 6, 7, 8, 9\}$ to evaluate model stability. As shown in Table~\ref{tab: window}, moderate session window sizes significantly improve performance by capturing useful historical context, while overly large session windows introduce noise and dilute current preferences. This result further confirms the stability of MISApp and its ability to effectively exploit relevant historical information.

\paragraph{Effectiveness of Transformer Layers}
We investigate how the number of Transformer layers affects performance by varying $L \in \{1, 2, 3, 4\}$. Performance improves from $L=1$ to $L=2$, whereas deeper architectures lead to diminishing returns or even overfitting. This suggests that a moderately deep architecture with $L=2$ provides an effective balance between expressive capacity and generalization ability for user intent modeling at the session-level. The experimental results are reported in Table~\ref{tab: Transformer Layers}.

\begin{table}[H]
\centering
\setlength{\tabcolsep}{2.5pt}
\renewcommand{\arraystretch}{1.0}
\caption{Effect of Maximum Session Window Size $T$ on Model Performance.}
\label{tab: window}
\begin{tabular}{c|ccccc}
\toprule
$T$ & ACC@1 & ACC@3 & ACC@5 & MRR@3 & MRR@5 \\
\midrule
5 & 53.39$_{\pm0.183}$ & 75.53$_{\pm0.251}$ & 81.07$_{\pm0.317}$ & 63.25$_{\pm0.209}$ & 64.67$_{\pm0.225}$ \\
6 & 53.36$_{\pm0.163}$ & 75.69$_{\pm0.318}$ & 82.26$_{\pm0.311}$ & 63.31$_{\pm0.224}$ & 65.43$_{\pm0.221}$ \\
7 & 53.87$_{\pm0.421}$  & 76.14$_{\pm0.724}$  & 82.29$_{\pm0.847}$  & 63.61$_{\pm0.542}$  & 65.54$_{\pm0.570}$  \\
8 &  \textbf{54.52$_{\pm0.175}$} & \textbf{76.33$_{\pm0.271}$} & \textbf{82.40$_{\pm0.240}$} & \textbf{64.26$_{\pm0.214}$} & \textbf{65.89$_{\pm0.209}$} \\
9 & 54.10$_{\pm0.109}$ & 75.75$_{\pm0.109}$ & 82.27$_{\pm0.126}$ & 63.79$_{\pm0.108}$ & 65.34$_{\pm0.111}$ \\
\bottomrule
\end{tabular}
\end{table}

\begin{table}[H]
\centering
\setlength{\tabcolsep}{2.5pt}
\renewcommand{\arraystretch}{1.0}
\caption{Effect of Transformer Layers $L$ on Model Performance.}
\label{tab: Transformer Layers}
\begin{tabular}{c|ccccc}
\toprule
$L$ & ACC@1 & ACC@3 & ACC@5 & MRR@3 & MRR@5 \\
\midrule
1 & 52.81$_{\pm0.034}$ & 73.65$_{\pm0.069}$ & 81.34$_{\pm0.055}$ & 62.71$_{\pm0.046}$ & 64.24$_{\pm0.042}$ \\
2 & \textbf{54.52$_{\pm0.175}$} & \textbf{76.33$_{\pm0.271}$} & \textbf{82.40$_{\pm0.240}$} & \textbf{64.26$_{\pm0.214}$} & \textbf{65.89$_{\pm0.209}$}\\
3 & 53.79$_{\pm0.032}$ & 74.76$_{\pm0.036}$ & 81.42$_{\pm0.051}$ & 63.79$_{\pm0.034}$ & 65.12$_{\pm0.036}$ \\
4 & 53.97$_{\pm0.109}$ & 74.55$_{\pm0.116}$ & 81.18$_{\pm0.144}$ & 63.63$_{\pm0.108}$ & 65.16$_{\pm0.113}$ \\
\bottomrule
\end{tabular}
\end{table}




\bibliographystyle{IEEEtran}
\bibliography{references}

@String{Computing = "Computing" }

@String{Computer = "{IEEE} Computer" }

@String{Springer = "Springer-Verlag" }

@article{ouyang2018modeling,
  title={Modeling and forecasting the popularity evolution of mobile apps: A multivariate hawkes process approach},
  author={Ouyang, Yi and Guo, Bin and Guo, Tong and Cao, Longbing and Yu, Zhiwen},
  journal={Proceedings of the ACM on Interactive, Mobile, Wearable and Ubiquitous Technologies},
  volume={2},
  number={4},
  pages={1--23},
  year={2018},
  publisher={ACM New York, NY, USA}
}

@inproceedings{c2,
  author    = {Li, Tong and Zhang, Mingyang and Cao, Hancheng and Li, Yong and Tarkoma, Sasu and Hui, Pan},
  title     = {``What Apps Did You Use?'': Understanding the Long-term Evolution of Mobile App Usage},
  booktitle = {Proc. Web Conf., Taipei, Taiwan},
  year      = {2020},
  pages     = {66--76}
}

@article{chen2019cap,
  title={CAP: Context-aware app usage prediction with heterogeneous graph embedding},
  author={Chen, Xinlei and Wang, Yu and He, Jiayou and Pan, Shijia and Li, Yong and Zhang, Pei},
  journal={Proceedings of the ACM on Interactive, Mobile, Wearable and Ubiquitous Technologies},
  volume={3},
  number={1},
  pages={1--25},
  year={2019},
  publisher={ACM New York, NY, USA}
}

@article{zhang2020app,
  title={App popularity prediction by incorporating time-varying hierarchical interactions},
  author={Zhang, Yixuan and Liu, Jiaqi and Guo, Bin and Wang, Zhu and Liang, Yunji and Yu, Zhiwen},
  journal={IEEE Transactions on Mobile Computing},
  volume={21},
  number={5},
  pages={1566--1579},
  year={2020},
  publisher={IEEE}
}

@article{zhu2013mobile,
  title={Mobile app classification with enriched contextual information},
  author={Zhu, Hengshu and Chen, Enhong and Xiong, Hui and Cao, Huanhuan and Tian, Jilei},
  journal={IEEE Transactions on mobile computing},
  volume={13},
  number={7},
  pages={1550--1563},
  year={2013},
  publisher={IEEE}
}

@article{li2025transformer,
  title={TGT: A Temporal Gating Transformer for Smartphone App Usage Prediction},
  author={Li, Longlong and Qu, Cunquan and Wang, Guanghui},
  journal={arXiv preprint arXiv:2502.16957},
  year={2025}
}

@inproceedings{liu2017effective,
author = {Liu, Junming and Fu, Yanjie and Ming, Jingci and Ren, Yong and Sun, Leilei and Xiong, Hui},
title = {Effective and Real-time In-App Activity Analysis in Encrypted Internet Traffic Streams},
year = {2017},
isbn = {9781450348874},
publisher = {Association for Computing Machinery},
address = {New York, NY, USA},
doi = {10.1145/3097983.3098049},
booktitle = {Proceedings of the 23rd ACM SIGKDD International Conference on Knowledge Discovery and Data Mining},
pages = {335–344},
numpages = {10},
location = {Halifax, NS, Canada},
series = {KDD '17}
}

@INPROCEEDINGS{9831540,
  author={Lu, Enze and Zhang, Long},
  booktitle={2022 IEEE 31st International Symposium on Industrial Electronics (ISIE)}, 
  title={Machine Learning Methods for Smartphone Application Prediction}, 
  year={2022},
  volume={},
  number={},
  pages={1174-1179},
  doi={10.1109/ISIE51582.2022.9831540}}

@article{li2022smartphone,
  title={Smartphone app usage analysis: Datasets, methods, and applications},
  author={Li, Tong and Xia, Tong and Wang, Huandong and Tu, Zhen and Tarkoma, Sasu and Han, Zhu and Hui, Pan},
  journal={IEEE communications surveys \& tutorials},
  volume={24},
  number={2},
  pages={937--966},
  year={2022},
  publisher={IEEE}
}

@inproceedings{chen2017powerful,
  author={Chen, Yimin and Jin, Xiaocong and Sun, Jingchao and Zhang, Rui and Zhang, Yanchao},
  booktitle={IEEE INFOCOM 2017 - IEEE Conference on Computer Communications}, 
  title={POWERFUL: Mobile app fingerprinting via power analysis}, 
  year={2017},
  volume={},
  number={},
  pages={1-9},
  address = { },
  location  ={Atlanta, GA, USA},
  doi={10.1109/INFOCOM.2017.8057232},
  publisher={IEEE}
  
}

@inproceedings{oliner2013carat,
author = {Oliner, Adam J. and Iyer, Anand P. and Stoica, Ion and Lagerspetz, Eemil and Tarkoma, Sasu},
title = {Carat: collaborative energy diagnosis for mobile devices},
year = {2013},
isbn = {9781450320276},
publisher = {Association for Computing Machinery},
address = {New York, NY, USA},
doi = {10.1145/2517351.2517354},
booktitle = {Proceedings of the 11th ACM Conference on Embedded Networked Sensor Systems},
articleno = {10},
numpages = {14},
location = {Roma, Italy},
series = {SenSys '13}
}

@article{xu2016understanding,
  title={Understanding mobile traffic patterns of large scale cellular towers in urban environment},
  author={Xu, Fengli and Li, Yong and Wang, Huandong and Zhang, Pengyu and Jin, Depeng},
  journal={IEEE/ACM transactions on networking},
  volume={25},
  number={2},
  pages={1147--1161},
  year={2016},
  publisher={IEEE}
}

@article{little_world,
  title={Collective dynamics of ‘small-world’networks},
  author={Watts, Duncan J and Strogatz, Steven H},
  journal={nature},
  volume={393},
  number={6684},
  pages={440--442},
  year={1998},
  publisher={Nature Publishing Group UK London}
}

@INPROCEEDINGS{zhao2019appusage2vec,
  author={Zhao, Sha and Luo, Zhiling and Jiang, Ziwen and Wang, Haiyan and Xu, Feng and Li, Shijian and Yin, Jianwei and Pan, Gang},
  booktitle={2019 IEEE 35th International Conference on Data Engineering (ICDE)}, 
  title={AppUsage2Vec: Modeling Smartphone App Usage for Prediction}, 
  year={2019},
  publisher = {IEEE},
  volume={},
  number={},
  pages={1322-1333},
  location = {Macao, China},
  doi={10.1109/ICDE.2019.00120}
}

@inproceedings{shin2012understanding,
author = {Shin, Choonsung and Hong, Jin-Hyuk and Dey, Anind K.},
title = {Understanding and prediction of mobile application usage for smart phones},
year = {2012},
isbn = {9781450312240},
publisher = {Association for Computing Machinery},
address = {New York, NY, USA},
doi = {10.1145/2370216.2370243},
booktitle = {Proceedings of the 2012 ACM Conference on Ubiquitous Computing},
pages = {173–182},
numpages = {10},
location = {Pittsburgh, Pennsylvania},
series = {UbiComp '12}
}

@inproceedings{baeza2015predicting,
author = {Baeza-Yates, Ricardo and Jiang, Di and Silvestri, Fabrizio and Harrison, Beverly},
title = {Predicting The Next App That You Are Going To Use},
year = {2015},
isbn = {9781450333177},
publisher = {Association for Computing Machinery},
address = {New York, NY, USA},
doi = {10.1145/2684822.2685302},
booktitle = {Proceedings of the Eighth ACM International Conference on Web Search and Data Mining},
pages = {285–294},
numpages = {10},
location = {Shanghai, China},
series = {WSDM '15}
}

@article{xia2020deepapp,
  title={DeepApp: Predicting personalized smartphone app usage via context-aware multi-task learning},
  author={Xia, Tong and Li, Yong and Feng, Jie and Jin, Depeng and Zhang, Qing and Luo, Hengliang and Liao, Qingmin},
  journal={ACM Transactions on Intelligent Systems and Technology (TIST)},
  volume={11},
  number={6},
  pages={1--12},
  year={2020},
  publisher={ACM New York, NY, USA}
}

@InProceedings{suleiman2021deeppatterns,
author="Suleiman, Basem
and Lu, Kevin
and Chan, Hong Wa
and Alibasa, Muhammad Johan",
editor="Hacid, Hakim
and Kao, Odej
and Mecella, Massimo
and Moha, Naouel
and Paik, Hye-young",
title="DeepPatterns: Predicting Mobile Apps Usage from Spatio-Temporal and Contextual Features",
booktitle="Service-Oriented Computing",
year="2021",
publisher="Springer International Publishing",
address="Cham",
pages="811--818",
isbn="978-3-030-91431-8"
}

@article{yu2020semantic,
  title={Semantic-aware spatio-temporal app usage representation via graph convolutional network},
  author={Yu, Yue and Xia, Tong and Wang, Huandong and Feng, Jie and Li, Yong},
  journal={Proceedings of the ACM on Interactive, Mobile, Wearable and Ubiquitous Technologies},
  volume={4},
  number={3},
  pages={1--24},
  year={2020},
  publisher={ACM New York, NY, USA}
}

@article{ouyang2022learning,
  title={Learning dynamic app usage graph for next mobile app recommendation},
  author={Ouyang, Yi and Guo, Bin and Wang, Qianru and Liang, Yunji and Yu, Zhiwen},
  journal={IEEE Transactions on mobile Computing},
  volume={22},
  number={8},
  pages={4742--4753},
  year={2022},
  publisher={IEEE}
}

@ARTICLE{huang2025predicting,
  author={Huang, Zihan and Li, Tong and Wang, Xing and Yang, Kexin and Deng, Chao and Feng, Junlan and Li, Yong},
  journal={IEEE Transactions on Mobile Computing}, 
  title={Predicting Mobile App Usage With Context-Aware Dynamic Hypergraphs}, 
  year={2025},
  volume={24},
  number={6},
  pages={5511-5524},
  doi={10.1109/TMC.2025.3532992}}

@article{khaokaew2024maple,
  title={Maple: Mobile app prediction leveraging large language model embeddings},
  author={Khaokaew, Yonchanok and Xue, Hao and Salim, Flora D},
  journal={Proceedings of the ACM on Interactive, Mobile, Wearable and Ubiquitous Technologies},
  volume={8},
  number={1},
  pages={1--25},
  year={2024},
  publisher={ACM New York, NY, USA}
}

@inproceedings{wang2020next,
author = {Wang, Jianling and Ding, Kaize and Hong, Liangjie and Liu, Huan and Caverlee, James},
title = {Next-item Recommendation with Sequential Hypergraphs},
year = {2020},
isbn = {9781450380164},
publisher = {Association for Computing Machinery},
address = {New York, NY, USA},
doi = {10.1145/3397271.3401133},
booktitle = {Proceedings of the 43rd International ACM SIGIR Conference on Research and Development in Information Retrieval},
pages = {1101–1110},
numpages = {10},
location = {Virtual Event, China},
series = {SIGIR '20}
}

@inproceedings{xu2019recurrent,
author = {Xu, Chengfeng and Zhao, Pengpeng and Liu, Yanchi and Xu, Jiajie and S.Sheng, Victor S.Sheng and Cui, Zhiming and Zhou, Xiaofang and Xiong, Hui},
title = {Recurrent Convolutional Neural Network for Sequential Recommendation},
year = {2019},
isbn = {9781450366748},
publisher = {Association for Computing Machinery},
address = {New York, NY, USA},
doi = {10.1145/3308558.3313408},
booktitle = {The World Wide Web Conference},
pages = {3398–3404},
numpages = {7},
location = {San Francisco, CA, USA},
series = {WWW '19}
}

@article{pareja2020evolvegcn, title={EvolveGCN: Evolving Graph Convolutional Networks for Dynamic Graphs}, volume={34}, DOI={10.1609/aaai.v34i04.5984}, abstractNote={&lt;p&gt;Graph representation learning resurges as a trending research subject owing to the widespread use of deep learning for Euclidean data, which inspire various creative designs of neural networks in the non-Euclidean domain, particularly graphs. With the success of these graph neural networks (GNN) in the static setting, we approach further practical scenarios where the graph dynamically evolves. Existing approaches typically resort to node embeddings and use a recurrent neural network (RNN, broadly speaking) to regulate the embeddings and learn the temporal dynamics. These methods require the knowledge of a node in the full time span (including both training and testing) and are less applicable to the frequent change of the node set. In some extreme scenarios, the node sets at different time steps may completely differ. To resolve this challenge, we propose EvolveGCN, which adapts the graph convolutional network (GCN) model along the temporal dimension without resorting to node embeddings. The proposed approach captures the dynamism of the graph sequence through using an RNN to evolve the GCN parameters. Two architectures are considered for the parameter evolution. We evaluate the proposed approach on tasks including link prediction, edge classification, and node classification. The experimental results indicate a generally higher performance of EvolveGCN compared with related approaches. The code is available at https://github.com/IBM/EvolveGCN.&lt;/p&gt;}, number={04}, journal={Proceedings of the AAAI Conference on Artificial Intelligence}, author={Pareja, Aldo and Domeniconi, Giacomo and Chen, Jie and Ma, Tengfei and Suzumura, Toyotaro and Kanezashi, Hiroki and Kaler, Tim and Schardl, Tao and Leiserson, Charles}, year={2020}, month={Apr.}, pages={5363-5370} 
}

@inproceedings{sankar2020dysat,
author = {Sankar, Aravind and Wu, Yanhong and Gou, Liang and Zhang, Wei and Yang, Hao},
title = {DySAT: Deep Neural Representation Learning on Dynamic Graphs via Self-Attention Networks},
year = {2020},
isbn = {9781450368223},
publisher = {Association for Computing Machinery},
address = {New York, NY, USA},
url = {https://doi.org/10.1145/3336191.3371845},
doi = {10.1145/3336191.3371845},
booktitle = {Proceedings of the 13th International Conference on Web Search and Data Mining},
pages = {519–527},
numpages = {9},
location = {Houston, TX, USA},
series = {WSDM '20}
}

@misc{xu2020inductive,
      title={Inductive Representation Learning on Temporal Graphs}, 
      author={Da Xu and Chuanwei Ruan and Evren Korpeoglu and Sushant Kumar and Kannan Achan},
      year={2020},
      eprint={2002.07962},
      archivePrefix={arXiv},
      primaryClass={cs.LG},
}

@inproceedings{vaswani2017attention,
 author = {Vaswani, Ashish and Shazeer, Noam and Parmar, Niki and Uszkoreit, Jakob and Jones, Llion and Gomez, Aidan N and Kaiser, {\L}ukasz and Polosukhin, Illia},
 booktitle = {Advances in Neural Information Processing Systems},
 editor = {I. Guyon and U. Von Luxburg and S. Bengio and H. Wallach and R. Fergus and S. Vishwanathan and R. Garnett},
 pages = {},
 publisher = {Curran Associates, Inc.},
 title = {Attention is All you Need},
 volume = {30},
 year = {2017}
}

@inproceedings{
kitaev2020reformer,
title={Reformer: The Efficient Transformer},
author={Nikita Kitaev and Lukasz Kaiser and Anselm Levskaya},
booktitle={International Conference on Learning Representations},
year={2020},
}

@inproceedings{wu2022timesnet,
  title={TimesNet: Temporal 2D-Variation Modeling for General Time Series Analysis},
  author={Haixu Wu and Tengge Hu and Yong Liu and Hang Zhou and Jianmin Wang and Mingsheng Long},
  booktitle={International Conference on Learning Representations},
  year={2023},
}

@InProceedings{zhou2022fedformer,
  title = 	 {{FED}former: Frequency Enhanced Decomposed Transformer for Long-term Series Forecasting},
  author =       {Zhou, Tian and Ma, Ziqing and Wen, Qingsong and Wang, Xue and Sun, Liang and Jin, Rong},
  booktitle = 	 {Proceedings of the 39th International Conference on Machine Learning},
  pages = 	 {27268--27286},
  year = 	 {2022},
  editor = 	 {Chaudhuri, Kamalika and Jegelka, Stefanie and Song, Le and Szepesvari, Csaba and Niu, Gang and Sabato, Sivan},
  volume = 	 {162},
  series = 	 {Proceedings of Machine Learning Research},
  month = 	 {17--23 Jul},
  publisher =    {PMLR}
}

@article{sun2025appformer,
  title={Appformer: A novel framework for mobile app usage prediction leveraging progressive multi-modal data fusion and feature extraction},
  author={Sun, Chuike and Chen, Junzhou and Zhao, Yue and Han, Hao and Jing, Ruihai and Tan, Guang and Wu, Di},
  journal={Expert Systems with Applications},
  volume={265},
  pages={125903},
  year={2025},
  publisher={Elsevier}
}

@inproceedings{he2020lightgcn,
author = {He, Xiangnan and Deng, Kuan and Wang, Xiang and Li, Yan and Zhang, YongDong and Wang, Meng},
title = {LightGCN: Simplifying and Powering Graph Convolution Network for Recommendation},
year = {2020},
isbn = {9781450380164},
publisher = {Association for Computing Machinery},
address = {New York, NY, USA},
doi = {10.1145/3397271.3401063},
booktitle = {Proceedings of the 43rd International ACM SIGIR Conference on Research and Development in Information Retrieval},
pages = {639–648},
numpages = {10},
location = {Virtual Event, China},
series = {SIGIR '20}
}

@article{li2023you,
  title={You are how you use apps: user profiling based on spatiotemporal app usage behavior},
  author={Li, Tong and Li, Yong and Zhang, Mingyang and Tarkoma, Sasu and Hui, Pan},
  journal={ACM Transactions on Intelligent Systems and Technology},
  volume={14},
  number={4},
  pages={1--21},
  year={2023},
  publisher={ACM New York, NY}
}

@ARTICLE{li2021finding,
  author={Li, Tong and Li, Yong and Xia, Tong and Hui, Pan},
  journal={IEEE Transactions on Network Science and Engineering}, 
  title={Finding Spatiotemporal Patterns of Mobile Application Usage}, 
  year={2021},
  volume={},
  number={},
  pages={1-1},
  doi={10.1109/TNSE.2021.3131194}
}

@article{yu2018smartphone,
  title={Smartphone app usage prediction using points of interest},
  author={Yu, Donghan and Li, Yong and Xu, Fengli and Zhang, Pengyu and Kostakos, Vassilis},
  journal={Proceedings of the ACM on Interactive, Mobile, Wearable and Ubiquitous Technologies},
  volume={1},
  number={4},
  pages={1--21},
  year={2018},
  publisher={ACM New York, NY, USA}
}

@article{aliannejadi2021context,
  title={Context-aware target apps selection and recommendation for enhancing personal mobile assistants},
  author={Aliannejadi, Mohammad and Zamani, Hamed and Crestani, Fabio and Croft, W Bruce},
  journal={ACM Transactions on Information Systems (TOIS)},
  volume={39},
  number={3},
  pages={1--30},
  year={2021},
  publisher={ACM New York, NY}
}

@article{zeng2023transformers, title={Are Transformers Effective for Time Series Forecasting?}, volume={37}, DOI={10.1609/aaai.v37i9.26317}, abstractNote={Recently, there has been a surge of Transformer-based solutions for the long-term time series forecasting (LTSF) task. Despite the growing performance over the past few years, we question the validity of this line of research in this work. Specifically, Transformers is arguably the most successful solution to extract the semantic correlations among the elements in a long sequence. However, in time series modeling, we are to extract the temporal relations in an ordered set of continuous points. While employing positional encoding and using tokens to embed sub-series in Transformers facilitate preserving some ordering information, the nature of the permutation-invariant self-attention mechanism inevitably results in temporal information loss. To validate our claim, we introduce a set of embarrassingly simple one-layer linear models named LTSF-Linear for comparison. Experimental results on nine real-life datasets show that LTSF-Linear surprisingly outperforms existing sophisticated Transformer-based LTSF models in all cases, and often by a large margin. Moreover, we conduct comprehensive empirical studies to explore the impacts of various design elements of LTSF models on their temporal relation extraction capability. We hope this surprising finding opens up new research directions for the LTSF task. We also advocate revisiting the validity of Transformer-based solutions for other time series analysis tasks (e.g., anomaly detection) in the future.}, number={9}, journal={Proceedings of the AAAI Conference on Artificial Intelligence}, author={Zeng, Ailing and Chen, Muxi and Zhang, Lei and Xu, Qiang}, year={2023}, month={Jun.}, pages={11121-11128} }

@article{chu2021air,
  title={Air pollution as a determinant of food delivery and related plastic waste},
  author={Chu, Junhong and Liu, Haoming and Salvo, Alberto},
  journal={Nature Human Behaviour},
  volume={5},
  number={2},
  pages={212--220},
  year={2021},
  publisher={Nature Publishing Group UK London}
}

@article{gruning2023directing,
  title={Directing smartphone use through the self-nudge app one sec},
  author={Gr{\"u}ning, David J and Riedel, Frederik and Lorenz-Spreen, Philipp},
  journal={Proceedings of the National Academy of Sciences},
  volume={120},
  number={8},
  pages={e2213114120},
  year={2023},
  publisher={National Academy of Sciences}
}

@inproceedings{khaokaew2021cosem,
author = {Khaokaew, Yonchanok and Rahaman, Mohammad Saiedur and White, Ryen W. and Salim, Flora D.},
title = {CoSEM: Contextual and Semantic Embedding for App Usage Prediction},
year = {2021},
isbn = {9781450384469},
publisher = {Association for Computing Machinery},
address = {New York, NY, USA},
doi = {10.1145/3459637.3482076},
booktitle = {Proceedings of the 30th ACM International Conference on Information \& Knowledge Management},
pages = {3137–3141},
numpages = {5},
location = {Virtual Event, Queensland, Australia},
series = {CIKM '21}
}

@Article{kmeans,
AUTHOR = {Ahmed, Mohiuddin and Seraj, Raihan and Islam, Syed Mohammed Shamsul},
TITLE = {The k-means Algorithm: A Comprehensive Survey and Performance Evaluation},
JOURNAL = {Electronics},
VOLUME = {9},
YEAR = {2020},
NUMBER = {8},
ARTICLE-NUMBER = {1295},
ISSN = {2079-9292},
DOI = {10.3390/electronics9081295}
}

@article{k-means++,
title = {Collaborative annealing power k-means++ clustering},
journal = {Knowledge-Based Systems},
volume = {255},
pages = {109593},
year = {2022},
issn = {0950-7051},
doi = {https://doi.org/10.1016/j.knosys.2022.109593},
author = {Hongzong Li and Jun Wang}
}

@article{kmodes,
  title={K-modes clustering algorithm for categorical data},
  author={Sharma, Neha and Gaud, Nirmal},
  journal={International Journal of Computer Applications},
  volume={127},
  number={17},
  pages={1--6},
  year={2015},
  publisher={Foundation of Computer Science}
}

@article{k-harmonic,
  title={Computer aided diagnostic system based on SVM and K harmonic mean based attribute weighting method},
  author={Srivastava, Anand Kumar and Kumar, Yugal and Singh, Pradeep Kumar},
  journal={Obesity Medicine},
  volume={19},
  pages={100270},
  year={2020},
  publisher={Elsevier}
}

\end{document}